%% file: main.tex
\documentclass[10pt,journal,compsoc]{IEEEtran}
\pdfoutput=1

\ifCLASSOPTIONcompsoc
  \usepackage[nocompress]{cite}
\else
  \usepackage{cite}
\fi

\usepackage{amsmath}
\usepackage{amssymb}
\usepackage{graphicx}
\usepackage{epsfig}
\usepackage{newfloat}
\usepackage{color}
\usepackage{adjustbox}
\usepackage{multirow}
\usepackage{verbatim}
\usepackage{tablefootnote}
\usepackage{framed}
\usepackage{url}
\usepackage{tikz}
\usepackage{tabulary,overpic,xcolor}
 \def\checkmark{\tikz\fill[scale=0.4](0,.35) -- (.25,0) -- (1,.7) -- (.25,.15) -- cycle;}
\newcommand{\app}{\raise.17ex\hbox{$\scriptstyle\sim$}}

\usepackage[caption=false]{subfig}
\newcommand{\newedit}[1]{#1}

\newcommand{\tablestyle}[2]{\setlength{\tabcolsep}{#1}\renewcommand{\arraystretch}{#2}\centering\footnotesize}
\newcolumntype{x}[1]{>{\centering\arraybackslash}p{#1pt}}

\begin{document}

\title{In the Eye of the Beholder: \\
Gaze and Actions in First Person Video}
% authors
\author{Yin Li,~\IEEEmembership{Member,~IEEE,}
        Miao Liu,~\IEEEmembership{Student Member,~IEEE,}
        James M.\ Rehg,~\IEEEmembership{Member,~IEEE,}
\IEEEcompsocitemizethanks{\IEEEcompsocthanksitem 
Y. Li is with the Department of Biostatistics and Medical Informatics and the Department of Computer Sciences, University of Wisconsin-Madison, 
Madison, WI, 53706. E-mail: yin.li@wisc.edu\protect\\
M. Liu and J.M. Rehg are with the School of Interactive Computing, College of Computing, Georgia Institute of Technology, 
Atlanta, GA, 30332. E-mail: mliu328@gatech.edu and rehg@gatech.edu\protect\\
}% <-this % stops a space
}% <-this % stops a space

% Extra Notes

% \markboth{IEEE TRANSACTIONS ON PATTERN ANALYSIS AND MACHINE INTELLIGENCE}%
% {Shell \MakeLowercase{\textit{et al.}}: In the Eye of the Beholder: Gaze and Actions in First Person Video}

% notes
\newcommand{\yin}[1]{\textcolor{blue}{yin: #1}}
\newcommand{\jim}[1]{\textcolor{red}{jim: #1}}
\newcommand{\miao}[1]{\textcolor{green}{miao: #1}}

\IEEEtitleabstractindextext{%
\begin{abstract}
We address the task of jointly determining what a person is doing and where they are looking based on the analysis of video captured by a headworn camera. To facilitate our research, we first introduce the EGTEA Gaze+ dataset. Our dataset comes with videos, gaze tracking data, hand masks and action annotations, thereby providing the most comprehensive benchmark for First Person Vision (FPV). Moving beyond the dataset, we propose a novel deep model for joint gaze estimation and action recognition in FPV. Our method describes the participant's gaze as a probabilistic variable and models its distribution using stochastic units in a deep network. We further sample from these stochastic units, generating an attention map to guide the aggregation of visual features for action recognition. Our method is evaluated on our EGTEA Gaze+ dataset and achieves a performance level that exceeds the state-of-the-art by a significant margin. More importantly, we demonstrate that our model can be applied to larger scale FPV dataset---EPIC-Kitchens even without using gaze, offering new state-of-the-art results on FPV action recognition. 
\end{abstract}

}

\maketitle
\IEEEdisplaynontitleabstractindextext
\IEEEpeerreviewmaketitle
\ifCLASSOPTIONcompsoc

\input{intro.tex}
\input{related_work.tex}
\input{dataset.tex}

\input{approach.tex}
\input{result.tex}

\input{conclusion.tex}

\ifCLASSOPTIONcompsoc
  \section*{Acknowledgments}
\else
  \section*{Acknowledgment}
\fi

This research was supported by grant U54EB020404 awarded by the National Institute of Biomedical Imaging and Bioengineering (NIBIB) through funds provided by the trans-NIH Big Data to Knowledge (BD2K) initiative \url{www.bd2k.nih.gov}. This work was also partially supported by Intel Science of Technology Center for Pervasive Computing (ISTC-PC). The work was developed during the first author's Ph.D.\ thesis at Georgia Tech.

\ifCLASSOPTIONcaptionsoff
  \newpage
\fi

% bibtex
\bibliographystyle{IEEEtran}
\bibliography{main}

% bio
\begin{IEEEbiography}[{\includegraphics[width=1in,height=1.25in,clip,keepaspectratio]{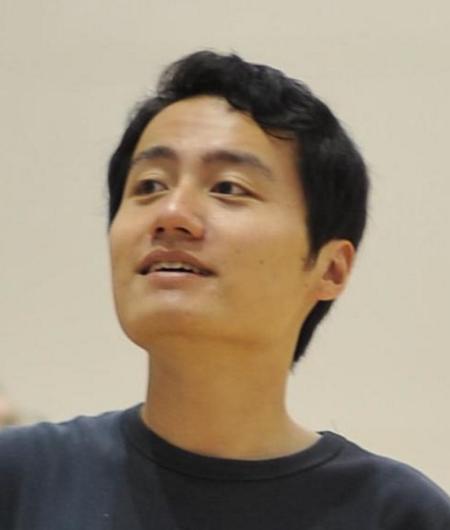}}]{Yin Li} is an Assistant Professor in the Department of Biostatistics and Medical Informatics and affiliate faculty in the Department of Computer Sciences at the University of Wisconsin-Madison. Previously, he obtained his PhD from the Georgia Institute of Technology and was a Postdoctoral Fellow at the Carnegie Mellon University. His primary research focus is computer vision. He is also interested in the applications of vision and learning for mobile health. He has been serving as area chairs for vision conferences CVPR, ICCV and ECCV. He is the co-recipient of the best student paper awards at MobiHealth 2014 and IEEE Face and Gesture 2015. 
\end{IEEEbiography}

\begin{IEEEbiography}[{\includegraphics[width=1in,height=1.25in,clip,keepaspectratio]{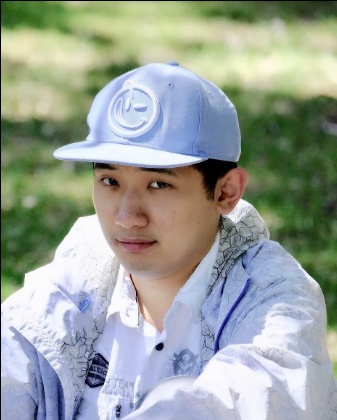}}]{Miao Liu} is currently a Robotics PhD student at Georgia Institute of Technology. His primary research focus is computer vision.  Especially, he is interested in understanding human attention and human actions from First Person Vision prospective.
\end{IEEEbiography}

\begin{IEEEbiography}[{\includegraphics[width=1in,height=1.25in,clip,keepaspectratio]{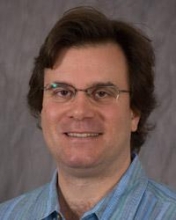}}]{James M. Rehg} is a Professor in the School of Interactive Computing at the Georgia Institute of Technology, where he is Director of the Center for Behavioral Imaging, co-Director of the Center for Computational Health, and co-Director of the Computational Perception Lab. He received his Ph.D. from CMU in 1995 and worked at the Cambridge Research Lab of DEC (and then Compaq) from 1995-2001, where he managed the computer vision research group. He received an NSF CAREER award in 2001 and a Raytheon Faculty Fellowship from Georgia Tech in 2005. He and his students have received a number of best paper awards, including best student paper awards at ICML 2005, BMVC 2010, Mobihealth 2014, Face and Gesture 2015, and a Method of the Year award from the journal Nature Methods. Dr. Rehg serves on the Editorial Board of the Intl. J. of Computer Vision, and he served as the General co-Chair for CVPR 2009 and the Program co-Chair for CVPR 2017. He has authored more than 100 peer-reviewed scientific papers and holds 23 issued US patents. Dr. Rehg's research interests include computer vision, machine learning, behavioral imaging, and mobile health (mHealth). 
\end{IEEEbiography}

\newpage

\end{document}

%% file: intro.tex
% Introduction
\IEEEraisesectionheading{\section{Introduction}\label{sec:introduction}}
\else
\section{Introduction}
\label{sec:introduction}
\fi

\IEEEPARstart{A}{dvances} in sensor miniaturization, low-power computing, and battery life have enabled the first generation of mainstream wearable cameras. Millions of hours of videos are captured by these devices every year, creating a record of our daily visual experience at an unprecedented scale. This has created a major opportunity to develop new capabilities and applications for computer vision. We have witnessed a surge of interest in the automatic analysis of visual data captured from wearable cameras, also known as First Person Vision (FPV)~\cite{kanade2012first} or Egocentric Vision. Examples include first person action and activity recognition~\cite{li2015delving,ma2016going,pirsiavash2012detecting,singh2016first,kitani2011fast,ryoo2013first,yonetani2016recognizing,su2016detecting}, first person gaze estimation and prediction~\cite{li2013learning,zhang2017deepgaze,soo2015social,park20123d}, first person pose estimation~\cite{rogez2015first,jiang2017pose} and first person video summarization~\cite{xu2015gaze,lee2015predicting,poleg2015egosampling}.

\begin{figure}[t]
\centering
\includegraphics[width=0.9\linewidth]{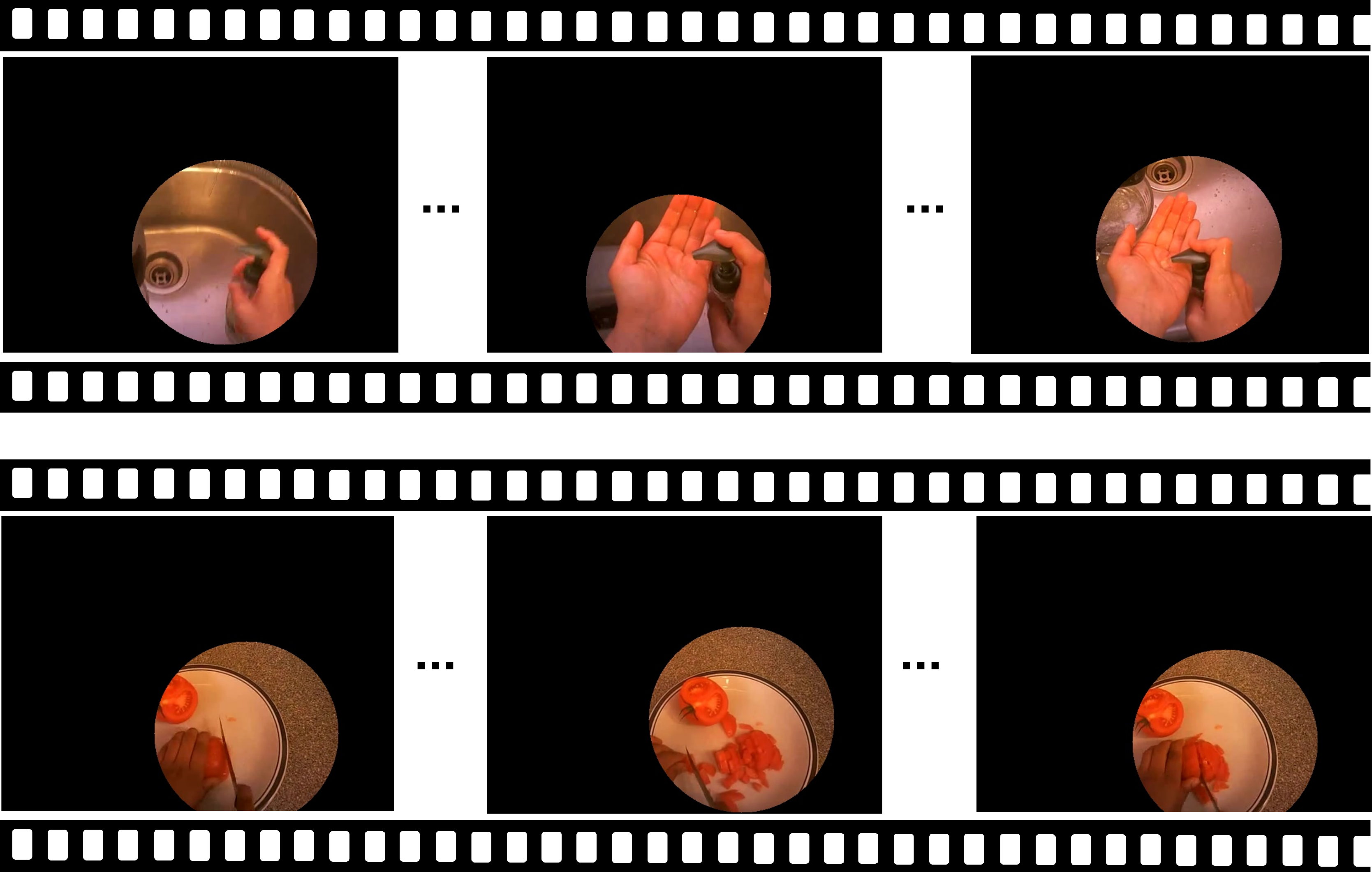}
\caption{{\it Can you tell what the person is doing? (examples taken from our EGTEA Gaze+ dataset)} With only $20\%$ of the pixels visible, centered around the point of gaze, we can easily recognize the camera wearer's actions. The gaze indexes key regions containing interactions with objects. We leverage this intuition and develop a model to jointly infer gaze and actions in First Person Vision.} \label{fig:teaser}\vspace{-1em}
\end{figure}

We address the problem of joint gaze estimation and action recognition in FPV. Our daily interaction with objects is guided by a sequence of carefully orchestrated fixations. It is thus critical to study the links between gaze and actions. We argue that ``where we look'' reveals important information about ``what we do.'' Consider the examples in Figure~\ref{fig:teaser}, where only small regions around the first person's point of gaze are shown. What is this person doing? We can easily identify the actions as ``squeeze liquid soap into hand'' and ``cut tomato," in spite of the fact that more than $80\%$ of the pixels are missing. This is possible because egocentric gaze serves as an index into the critical regions of the video that define the action. Focusing on these regions eliminates the potential distraction of irrelevant background pixels, and allows us to focus on the key elements of the action. In this case, attention is naturally embodied in the camera wearer's actions. Thus, FPV provides the ideal vehicle for studying the joint modeling of attention and action.

In order to study attention and action in FPV, we extend our previous work~\cite{fathi2012learning,li2015delving} and introduce the Extended GTEA Gaze+ (EGTEA Gaze+) dataset---a new FPV action dataset of meal preparation tasks captured in a naturalistic kitchen environment. Our work is related to recent efforts~\cite{Damen2018EPICKITCHENS} to create large scale benchmarks for FPV action recognition. In contrast to concurrent work, such as EPIC-Kitchens~\cite{Damen2018EPICKITCHENS} and Charades-Ego~\cite{sigurdsson2018charadesego}, our dataset not only includes FPV videos with fine-grained \emph{action annotations}, but also provides first person \emph{gaze tracking data} and egocentric \emph{hand masks}. We believe our EGTEA Gaze+ dataset offers the most comprehensive benchmark for FPV to date. Our dataset is publicly available\footnote{Our dataset is available at~\url{http://cbi.gatech.edu/fpv}} and has been considered by several recent works~\cite{sudhakaran2019lsta,furnari2019would,luo2019grouped,huang2019mutual,bandini2020analysis,liu2019forecasting}.

Moving beyond the dataset, a major challenge for the joint modeling of egocentric gaze and action is the uncertainty in gaze measurements. A significant portion of the egocentric gaze events are irrelevant to the actions. For instance, around $25\%$~\cite{henderson2003human} of our gaze within daily actions are saccades --- rapid gaze jumps during which our vision system receives no inputs~\cite{bridgeman1975failure}. Within the gaze events that remain, it is not clear what portion of the fixations correspond to overt attention and are therefore meaningfully-connected to actions~\cite{itti2001computational}. In addition, there are small but non-negligible measurement errors in the eye-tracker itself~\cite{hansen2010eye}. It follows that a joint model of attention and actions must account for the uncertainty of gaze. What model should we use to represent this uncertainty?

Our inspiration comes from the observation that gaze can be characterized by a \emph{latent} distribution of attention in the context of an action, represented as an attention map in egocentric coordinates. This map identifies image regions that are salient to the current action, such as hands, objects, and surfaces. We model gaze measurements as samples from the attention map distribution. Given gaze measurements obtained during the production of actions, we can directly learn a model for the attention map, which can in turn guide action recognition. Our action recognition model can then focus on action-relevant regions to determine what the person is doing. The attention model is tightly coupled with the recognition of actions. Building on this intuition, we develop a deep network with a latent variable attention model and an attention mechanism for recognition. 

To this end, we propose a novel deep model for joint gaze estimation and action recognition in FPV. Specifically, we model the latent distribution of gaze as stochastic units in a deep network. This representation allows us to sample attention maps from the gaze distribution. These maps are further used to selectively aggregate relevant visual features in space and time for action recognition. Our model thus encodes the uncertainty in gaze measurement, and models visual attention in the context of actions. When gaze measurement is available during training, we train the model in an end-to-end fashion using action labels and noisy gaze measurements as supervision. When gaze measurement is not available, we explore training our model using a simple prior distribution. At test time, our model receives only an input video and is able to infer both gaze and action.

We first evaluate our model on the new EGTEA Gaze+ dataset. As a consequence of jointly modeling gaze and actions, we obtain results for action recognition that outperform state-of-the-art deep models by a significant margin on EGTEA Gaze+. Our gaze estimation accuracy is also comparable with several strong baseline methods on EGTEA Gaze+. More importantly, we demonstrate that the model architecture developed on EGTEA Gaze+ can be effectively transferred to a larger scale FPV action dataset---EPIC-Kitchens. We develop a version of our model that does not require direct gaze supervision, yet uses a uniform prior of attention for training. Our joint model achieves state-of-the-art results on the challenging EPIC-Kitchens dataset.

We summarize our key contributions as follows: 
\begin{itemize}
\item We present the EGTEA Gaze+ dataset, the most comprehensive FPV dataset with gaze tracking data, annotated fine-grained actions and hand masks. We establish solid benchmark using the proposed dataset for FPV action recognition. We believe our dataset and benchmark will provide a major resource for the community. 
\item We propose a novel deep model for joint gaze estimation and action recognition in FPV. At the core of our model lies in the probabilistic modeling of visual attention using stochastic units in a deep network. To the best of our knowledge, this is the first work to model \emph{uncertainty} in gaze measurements for action recognition, and the first deep model for \emph{joint} gaze estimation and action recognition in FPV.  
\item Our method achieves state-of-the-art results on the new EGTEA Gaze+ dataset. More importantly, we demonstrate that our model can be applied to EPIC-Kitchens---the largest FPV dataset even without using gaze, leading to new state-of-the-art results. 
\end{itemize}

A preliminary version of this paper was discussed in the first author's thesis~\cite{li2017learning} and also appeared in ECCV'18~\cite{li2018eccv}. \newedit{This journal version extends our conference paper in several important aspects. First, we include a detailed description of the dataset effort that was not in our previous paper. Second, we improve our previous results by using additional regularization during training, report full set of results on all splits of EGTEA Gaze+ and compare them to latest methods. Third, we extend our previous model to handle training without ground-truth human gaze. This extension has enabled the application of our model on other FPV datasets like EPIC-Kitchens~\cite{Damen2018EPICKITCHENS}. Finally, we demonstrate that our model can generalize to EPIC-Kitchens dataset which does not have gaze tracking data. Our model achieves competitive results on EPIC-Kitchens Challenge 2020, ranked 2nd for unseen environments and 4th for seen environments.\footnote{See results on \url{https://epic-kitchens.github.io/2020-55.html}}}

Our paper is organized as follows. Section 2 covers related work on first person vision. Section 3 presents our dataset. Section 4 describes our model for joint gaze estimation and action recognition. Sections 5 details our experiments. Finally, Section 6 summarizes our findings and discusses future directions for FPV. 

%% file: related_work.tex
\section{Related Work}
\label{sec:related_work}

\subsection{Action Recognition}
We briefly review the literature on action recognition in computer vision. The main property of this work is that it assumes a third person view of one or more individuals, e.g., as would be captured by a surveillance camera, and asks what the individuals are doing. A thorough survey of this previous work is beyond our scope, and we refer the readers to recent survey papers~\cite{reviewUT,reviewUMD} for a comprehensive description. We discuss relevant work on the development of deep models and the use of attentional cues for recognizing actions.

\noindent \textbf{Deep Models for Action Recognition}. \newedit{Deep models have been immensely successful in action recognition. Early works used 2D convolutional networks and recurrent networks for action recognition~\cite{karpathy2014large,donahue2015long}. Simonyan and Zisserman~\cite{simonyan2014two} proposed a two-stream network to recognize actions from both optical flow maps and RGB video frames. Wang et al.\ \cite{wang2016temporal} further extended the two-stream network to model multiple temporal segments of a video. Feichtenhofer et al.\ \cite{feichtenhofer2016convolutional} explored different ways of fusing the RGB and flow streams. Features from these deep models can be combined with either tracked local descriptors~\cite{wang2015action} or human poses estimation results~\cite{cheron2015Pose} for action understanding. 

Going forward, Du et al.\ \cite{tran2014learning} replaced 2D convolutions with spatial temporal convolutions and proposed to learn a 3D convolutional neural network for action recognition. Carreira and Zisserman further proposed a two-stream 3D convolutional network for action recognition~\cite{carreira2017quo}. Several extensions of 3D convolutional networks have since been explored, including mixed 2D and 3D networks~\cite{Tran_2018_CVPR, Xie_2018_ECCV}, grouped or depthwise 3D convolutions~\cite{hara2018can,chen2018multi,luo2019grouped,tran2019video}, and the use of network architecture search~\cite{piergiovanni2019evolving}. Build on two-stream 3D networks, Feichtenhofer et al.\ \cite{feichtenhofer2019slowfast} proposed a SlowFast network, combining a low frame rate, high resolution slow stream, and a high frame rate, low resolution fast stream. More recent works focused on learning video representations~\cite{misra2016shuffle,lee2017unsupervised,owens2018audio,wei2018learning,crasto2019mars,sun2019videobert} (including ours~\cite{liu2019paying}), and the modeling of long-range dependencies in videos~\cite{varol2017long, wang2018none,wu2019long}. Our model builds on the success of 3D convolutional networks~\cite{carreira2017quo,tran2019video} to recognize actions in FPV. Our key technical novelty is to incorporate stochastic units for the joint modeling of egocentric gaze and actions.}

\noindent \textbf{Attention for Actions}. Human gaze provides useful signals for the location of actions. This intuition has been explored for action recognition in domains outside of FPV. \newedit{For example, several previous works on actionness map~\cite{wang2016actionness,chen2014actionness} aimed at estimating regions of actions from raw video frames.} Moreover, Mathe and Sminchesescu~\cite{mathe} proposed to recognize actions by sampling local descriptors from a predicted saliency map. Shapovalova et al.\ \cite{gregNIPS} presented a method that uses human gaze for learning to localize actions. However, these methods did not use deep models. Recently, Shikhar et al.\ \cite{sharma2015action} incorporated soft attention into a deep recurrent network for recognizing actions. However, their notion of attention is defined by discriminative image regions that are not derived from gaze measurements, and thus these previous approaches can't support the joint inference of egocentric gaze and actions.

Our method shares a key intuition as~\cite{mathe,gregNIPS} by using predicted gaze to select visual features. Unlike~\cite{mathe,gregNIPS}, our attention model is integrated within a deep network and trained from end-to-end. Our model is also similar to~\cite{sharma2015action} as we also design a attention mechanism that facilitates end-to-end training. Unlike~\cite{sharma2015action}, attention is modeled as stochastic units in our model and receives supervision from either noisy human gaze measurements or a prior distribution. 

\noindent \textbf{Datasets for Action Recognition}. A major driving force behind the recent advance in action recognition is the development of large-scale video datasets and benchmarks. Examples include UCF101~\cite{soomro2012ucf101},  HMDB~\cite{kuehne2011hmdb} and more recent 20BN-Something-Something~\cite{goyal2017something}, Charades~\cite{sigurdsson2016hollywood} and Kinetics~\cite{carreira2017quo}, where tens of thousands of video clips were collected from Internet and annotated manually. In the meanwhile, previous FPV action datasets, including our own work~\cite{fathi2012learning}, lags behind in terms of number of samples. Similar to the concurrent work of EPIC-Kitchens~\cite{Damen2018EPICKITCHENS} and Charades-Ego~\cite{sigurdsson2018charadesego}, our work seeks to bridge this gap. Different from existing efforts, our key focus is to provide a more comprehensive set of signals for attention and action in FPV. Specifically, EGTEA Gaze+ dataset includes FPV videos, gaze tracking, action annotations and hand masks. We will describe our effort on EGTEA Gaze+ in Sec.\ \ref{sec:dataset}.

Another highly relevant dataset is the MPII-Cooking dataset~\cite{rohrbach2012database}. Both datasets focus on cooking activities, with MPII following a conventional 3rd person paradigm. Our dataset, in contrast, was captured from the first person perspective, and it offers the largest benchmark for FPV action recognition, gaze estimation and hand segmentation. Our dataset is also related to the ADL dataset from Pirsiavash and Ramanan~\cite{pirsiavash2012detecting}, where they collected and annotated 10 hours of FPV videos. However, ADL is targeted for complex activities (Activities of Daily Living) and is substantially smaller than our dataset in terms of number of instances. We believe that our EGTEA Gaze+ dataset can serve as a major resource for the community to further advance the understanding of attention and actions in FPV.

\subsection{First Person Vision}
We now describe the emerging field of first person vision and its related work. In this section, we focus on action and activity recognition in FPV. Other efforts include egocentric gaze estimation~\cite{li2013learning,park20123d,soo2015social}, hand analysis~\cite{cmuHand,huang2015how,rogez2015understanding,feix2016grasp,Tekin2019CVPR}, pose estimation~\cite{rogez2015first,jiang2017pose}, physiological parameter estimation~\cite{hernandez2014bioglass,Nakamura2017egoheart}, user identification~\cite{hoshen2016egocentric,poleg2014head,Yonetani2015egosurfing} and video summarization~\cite{xu2015gaze,lee2015predicting,poleg2015egosampling}. A recent survey of this literature can be found in~\cite{betancourt2015evolution}. 

\noindent \textbf{FPV Gaze}. Gaze estimation is well studied in computer vision~\cite{borji2013state}. Several recent work have addressed the problem of egocentric gaze estimation. Our previous work \cite{li2013learning} estimated egocentric gaze using hand and head cues. Zhang et al.\ \cite{zhang2017deep} predicted future gaze by estimating gaze from predicted future frames. Huang et al.\ \cite{huang2018predicting} modeled the transition of attention for FPV gaze estimation. Park et al.\ \cite{park20123d} considered $3$D social gaze from multiple camera wearers. However, these work did not model egocentric gaze in the context of actions.

\noindent \textbf{FPV Actions}. FPV action has been the subject of many recent efforts. Spriggs et al.\ \cite{spriggs2009temporal} proposed to segment and recognize daily activities using a combination of video and wearable sensor data. Kitani et al.\ \cite{kitani2011fast} used a global motion descriptor to discover egocentric actions. Fathi et al.\ \cite{fathi2011understanding} presented a joint model of objects, actions and activities. Pirsiavash and Ramanan~\cite{pirsiavash2012detecting} further advocated for an object-centric representation of FPV activities. Other efforts included the modeling of conversations~\cite{fathi2012social} and reactions~\cite{yonetani2016recognizing,Li2019CVPR} in social interactions. 

Several recent work have developed deep models for FPV action recognition. Ryoo et al.\ \cite{ryoo2015pooled} developed a novel pooling method for deep models. Poleg et al.\ \cite{poleg2016compact} used temporal convolutions on motion fields for long-term activity recognition. Kazakos et al.\ \cite{kazakos2019epic} proposed to fuse video and audio signals for FPV action recognition. Wray et al.\ \cite{wray2019fine} made use of text descriptions of FPV actions for zero-short learning. In contrast to our work, these prior work did not consider using egocentric gaze for action recognition. 

\noindent \textbf{FPV Gaze and Actions}. There have been a few work that incorporated egocentric gaze for FPV action recognition. \newedit{For example, our previous work~\cite{li2015delving} showed the benefits of gaze-indexed or hand-indexed visual features for FPV action recognition.} Both Singh et al.\ \cite{singh2016first} and Ma et al.\ \cite{ma2016going} explored the use of multi-stream networks to capture egocentric attention. Moreover, Shen et al.\ \cite{shen2018egocentric} modeled gaze event towards objects for future egocentric action prediction. These works have clearly demonstrated the advantage of using egocentric gaze for FPV actions. However, they all model FPV gaze and actions \emph{separately} rather than jointly, and they do not address the uncertainty in gaze. Moreover, these methods require \emph{side information} in addition to the input image at testing time, e.g., hand masks~\cite{singh2016first,li2015delving}, object information~\cite{ma2016going}, or human gaze~\cite{shen2018egocentric}. 

\newedit{More recently, Sudhakaran et al.\ \cite{sudhakaran2018attention, sudhakaran2019lsta} presented LSTM models with soft attention for FPV action recognition. Furnari and Farinella~\cite{furnari2019would} proposed to combine two LSTMs with attention for FPV action anticipation. However, these methods did not consider human gaze. In contrast, our method presents \emph{the first joint model of gaze and action in FPV}, captures the uncertainty of gaze, and requires only video inputs during testing. Subsequent to our work, Huang et al.\ \cite{huang2019mutual} revisited the joint modeling and proposed to further model the top-down modulation of gaze given actions. Kapidis et al.\ \cite{kapidis2019multitask} presented a multi-stream networks that learns to output gaze, hand location, objects (noun) and motion (verb) at the same time.}

Our previous work~\cite{fathi2012learning} presented a joint model for FPV gaze and actions. This work extends~\cite{fathi2012learning} in multiple aspects: (1) we propose an end-to-end deep model rather than using hand crafted features; (2) we explicitly model ``noise'' in gaze measurements while~\cite{fathi2012learning} did not; (3) we infer gaze and action jointly through a single pass during testing while~\cite{fathi2012learning} used iterative inference. In a nutshell, we model gaze as a stochastic variable via a novel deep architecture for joint gaze estimation and action recognition. Our model thus combines the benefits of latent variable modeling with the expressive power of a learned feature representation. Consequently, we show that our method can outperform latest deep models~\cite{carreira2017quo} for FPV action recognition. \newedit{Our recent work~\cite{liu2019forecasting} has further developed this idea to model future hand trajectories for FPV action anticipation.}

%% file: dataset.tex
\section{The Extended GTEA Gaze+ Dataset}
\label{sec:dataset}

We start by presenting our work on creating EGTEA Gaze+ dataset---a major expansion of our previous GTEA Gaze+ dataset~\cite{fathi2012learning,li2015delving}. Specially, our new dataset contains more than $28$ hours of videos, $3$ times larger than GTEA Gaze+. Theses videos are from $86$ unique sessions of $32$ subjects performing $7$ different meal preparation tasks. Our new dataset subsumes GTEA Gaze+ as a subset, yet with revised annotations via our new annotation pipeline. The final dataset comes with videos, gaze tracking, action annotations and hand masks. We hope our dataset will serve as a major vehicle for understanding first person gaze and actions. We now describe our data collection and annotation process.

\subsection{Data Collection}
\begin{figure}[t]
\centering
\includegraphics[width=0.95\linewidth]{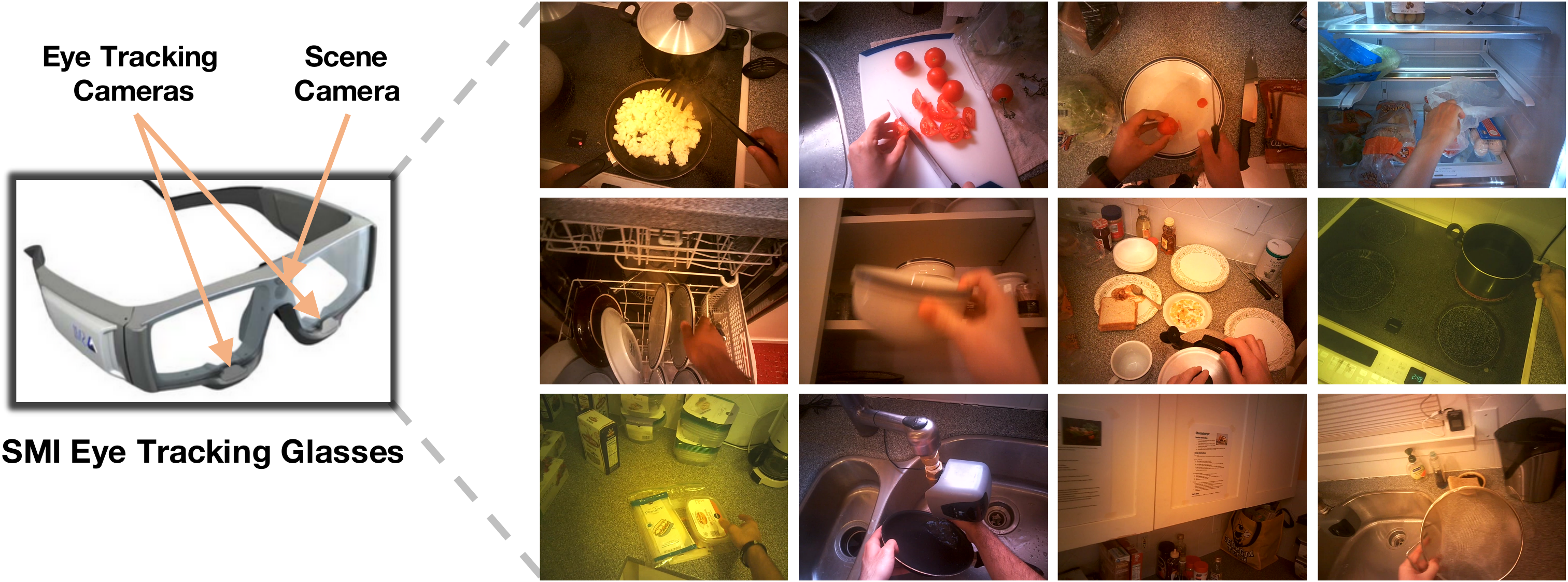}
\caption{Left: SMI eye tracking glasses used for video recording. Right: sample frames from the videos. Our dataset contains videos with different lighting condition, object instances and actions.}\label{fig:extended_gp}\vspace{-1em}
\end{figure}

The dataset was collected at the kitchen area of Aware Home\footnote{\url{http://www.awarehome.gatech.edu/}} on Georgia Tech campus. This kitchen area provides a naturalistic house-holding environment that contains the standard appliances, furnishings and food. All participants were recruited from Georgia Tech. A written consent was obtained for each participant, such that their FPV recording can be used and shared for research purpose. The study was approved by Institutional Review Board (IRB). We recorded videos, audios and gaze tracking data using the SMI eye-tracking glasses\footnote{ \url{https://imotions.com/hardware/smi-eye-tracking-glasses/}} (see Figure~\ref{fig:extended_gp} (left)). We now describe our study protocol and how we de-identify the videos. 

\noindent \textbf{Protocol}. At the beginning of the session, a researcher introduced the protocol, present a target recipe\footnote{7 recipes were considered, including breakfast (scrambled eggs), snack (peanut-butter sandwich), turkey sandwich, pizza, Greek salad, pasta salad and cheese burger.} to the participants and answered any questions they may have. Each recipe included detailed key steps of the dishes, such as ``fill a small pot with water''. The participants were given a few minutes to explore the kitchen and to go through the recipe. The same researcher then helped the participant to wear the eye-tracking glasses that is connected to a host laptop. A calibration of the eye tracker was first performed by asking the participant to stand still and look at a few landmarks in the kitchen. After the calibration, the researcher helped the participant to wear a backpack with the host laptop inside. The participant was then able to move freely within the kitchen, with FPV video and gaze captured by the glasses and stored on the laptop in the backpack. \newedit{The laptop and the backpack weighted under 2 kilograms in total. Based on our observation, this weight had minimum impact of the participants' behavior, including both gaze and actions.}

During the session, the participant was asked to prepare the dish by following the recipe. Paper copies of the recipe were available on site and the participant was free to check the recipe during a session. No other instruction were given. The participant can choose to end anytime during the session, yet the majority of them ended the session after they finished the dish. At the end of a session, another calibration process for the eye tracker was performed similar to the one at the beginning of the session. And the eye tracking quality was manually verified by the researcher. Sessions with low tracking quality were discarded from our dataset. Finally, the researcher helped the participant to remove the glasses. 

Each session usually lasts between 10 minutes and half an hour. Across different sessions, we altered the lighting condition by turning on different light blobs or putting down the blinds of the window. The object instances also varied as we re-supplied the utensils and food. Our goal is to have videos that exhibit a high variety of lighting conditions, objects and actions.

\begin{figure}[t]
\centering
\includegraphics[width=0.9\linewidth]{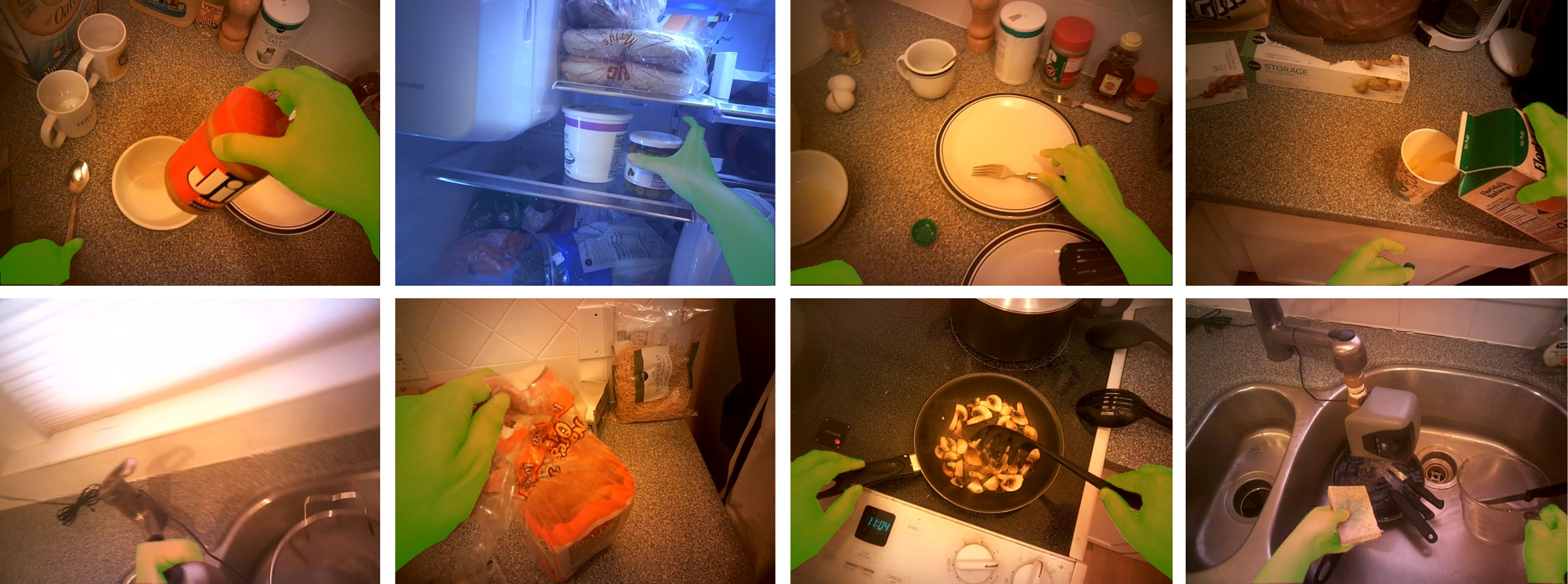}
\caption{Ground truth hand masks from our Hand14K dataset. The annotated masks are shown as green regions.}\label{fig:hand_gt}\vspace{-1em}
\end{figure}

\noindent \textbf{Video De-identification}. We manually screened all recorded videos to remove any frames that might reveal the identity of the participant. Most of the removed frames are around the start and end of a session, when the participant was mounting or removing the glasses. In rare case, we removed frames that have reflections of the participant's face in the environment, e.g., reflections on a mirror-finished stainless kettle. The screened video was further verified by another researcher. After the screening, more than $95\%$ of all frames were kept. All removed frames were replaced with an empty black image in the videos.

\noindent \textbf{Video Data Statistics}. We recruited 32 participants across 7 recipes. At the end, we obtained 86 videos (sessions) with high quality eye tracking and a total of 28 hours. Each of the video has a resolution of $1280\times 920$ captured at $24$Hz with a field of view (FoV) of $60^\circ \times 46^\circ$ (horizontal $\times$ vertical) . All videos come with audio and gaze tracking data. Sample frames of the videos are shown in Figure~\ref{fig:extended_gp} (right).

\begin{figure}[t]
\centering
\includegraphics[width=0.9\linewidth]{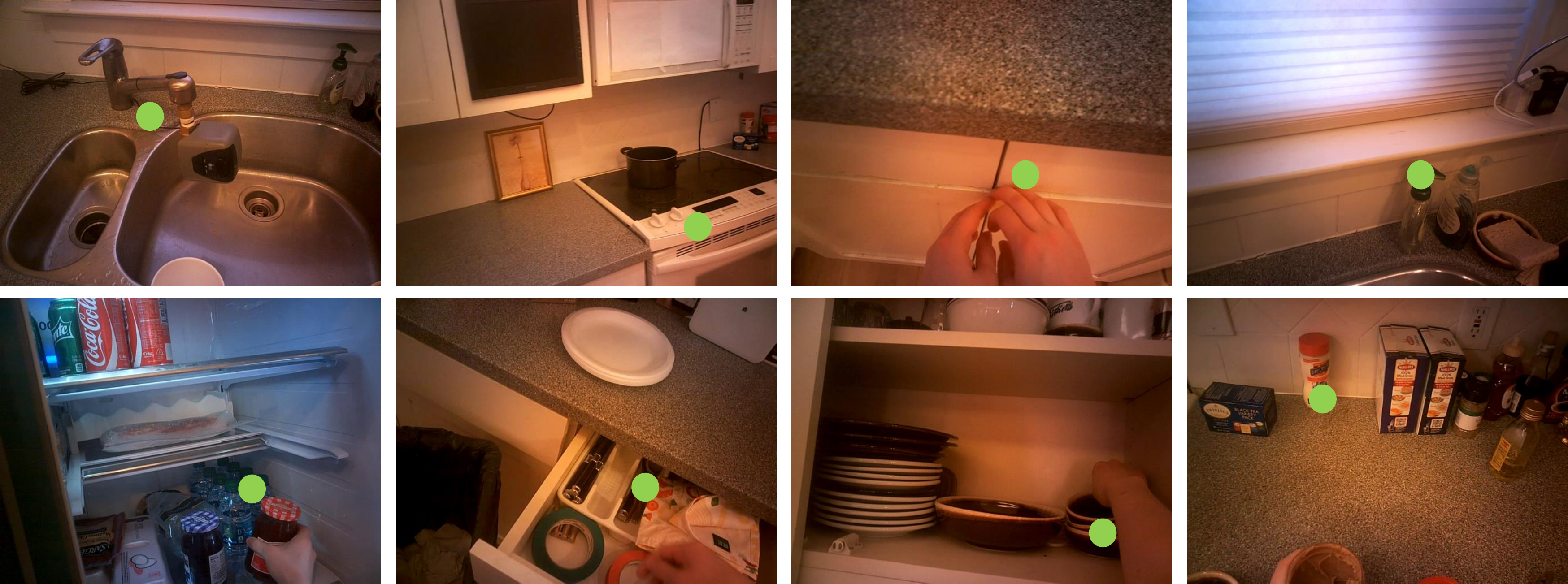}
\caption{Gaze tracking data from our dataset. The tracked gaze points are shown as green dots on the video frames.}\label{fig:gaze_sample}\vspace{-1em}
\end{figure}

\noindent \textbf{Gaze Data Statistics}. Binocular gaze data was tracked at 30Hz and synchronized with the recorded videos by SMI mobile eye tracker. Each gaze point is time-stamped and defined as a 2D point in the image plane. Furthermore, a proprietary software from SMI was used to identify the tracked gaze into (1) a saccade (rapid eye movement), (2) a fixation or (3) an unknown gaze type. In very rare cases (3.3\%), the gaze tracker might fail to track the egocentric gaze (untracked). Over a total of 3-million gaze points, the ratio of fixation, saccade, unknown gaze type and untracked gaze are 53.6\%, 26.1\%, 17.0\% and 3.3\%, respectively. Sample gaze points are shown in Figure~\ref{fig:gaze_sample}.

\subsection{Data Annotation}
We further annotated the dataset by pixel-level annotations of the first person's hands for sparsely sampled frames, as well as frame-level annotations of actions for all videos. We present our annotation details. 

\subsubsection{Hand Mask Annotation}
In addition to action actions, we provide egocentric hand mask annotations on sparsely sampled video frames. Egocentric hands are important cues for FPV actions, as we use our hands to interact with the objects and the physical environment. Specifically, we sampled one frame from every 5 seconds within the video. Empty (due to de-identification) or blurry frames were removed. And the rest of the frames were out-sourced to a third-party company, where a modified interface from~\cite{bell2013opensurfaces} was used to contour the egocentric hands. Each hand was annotated using one or more polygons (if it is been occluded) following its contour, which was further converted into a hand mask. After the annotation, we went through the hand masks and further removed poorly annotated one.

\noindent \textbf{Hand Mask Statistics}. We obtained a total of number of $15,176$ hand masks from $13,847$ frames. Both frames and the hand masks are in the resolution of $960\times 720$. On average, there are $1.1$ masks per image, and each image has $6.0\%$ of hand pixels. We released this dataset as our Hand14K dataset, an important part of the EGTEA Gaze+ dataset. Figure~\ref{fig:hand_gt} shows sample annotations of hand masks from Hand14k. We hope this dataset will provide a major resource for analyzing hands in FPV.

\subsubsection{Action Annotation}
To capture FPV actions, we use the same taxonomy from~\cite{sigurdsson2017new} to define action labels. Moreover, we develop a multi-stage pipeline to enable accurate and efficient video annotation. We discuss the action categories on our dataset, present our pipeline and analyze the result action labels.

\noindent \textbf{Action Categories}. Our first step is to identify the action categories. In this work, we focus on fine-grained actions that can be described by a combination of a single verb and a set nouns, such as ``take tomato'' or ``turn on oven''. The verb describes the motion, e.g., ``take'', ``turn on'', and the nouns specify the objects involved in the action, e.g., ``tomato'' or ``peanut butter container''. We did not distinguish between the plural or singular form of the nouns, and thus instead focus on the recognizing the presence of the objects. This combination of verb and nouns can describe complex actions, such as ``pour condiment (from) container (to) salad''. A similar naming taxonomy is also used in~\cite{sigurdsson2016hollywood,Damen2018EPICKITCHENS} and discussed in~\cite{sigurdsson2017new}.

\begin{figure*}[t]
\centering
\includegraphics[width=0.75\linewidth]{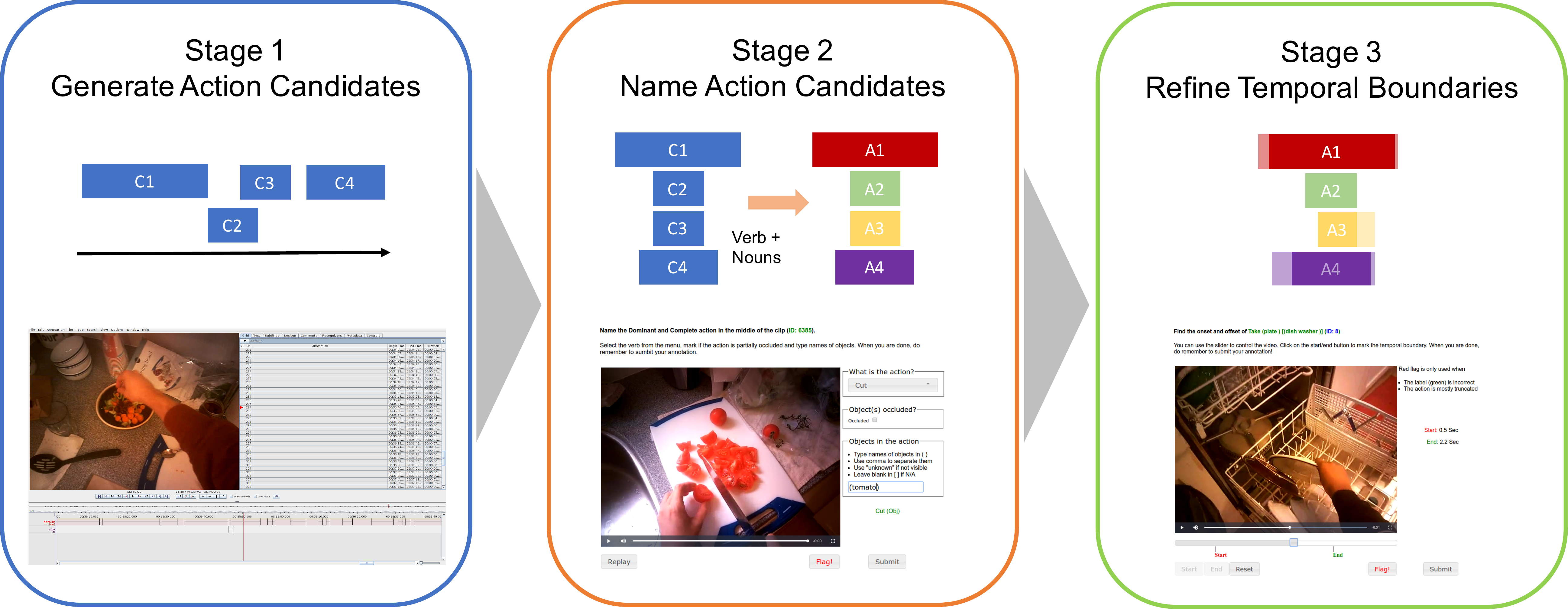}
\caption{Action annotation pipeline. We follow a three stage pipeline for annotation, with each stage focuses on a single task. From left to right: interfaces for action candidate labeling, action naming and action trimming. We use ELAN~\cite{wittenburg2006elan} for generating action candidates---clips that contain full extent of an action. Moreover, we developed an interactive web User Interface (UI) for further label the clips (action naming) and refine the temporal boundary of the action (action trimming).}\label{fig:annotation_pipeline}
\end{figure*}

\begin{figure*}[ht]
\centering
\includegraphics[width=0.9\linewidth]{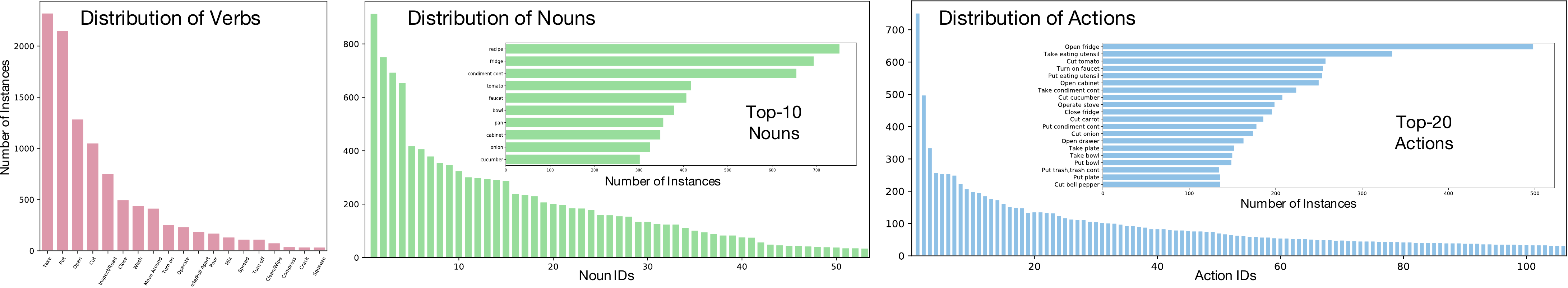}
\caption{Long tailed distribution of verbs (left), nouns (middle) and actions (right) in our dataset. We consider $19$ verbs, $53$ nouns and $106$ fine-grained actions. Top-10 objects and top-20 actions are further displayed. The distribution poses additional challenge of learning from imbalanced data.}\label{fig:new_data_stats}\vspace{-1em}
\end{figure*}

\noindent \textbf{Annotation Pipeline}. Annotating an action in a continuous video requires to identify its onset and offset, and to generate its label according to our taxonomy. This process can be very time consuming for even an expert human annotator. We propose to streamline the annotation by dividing the process into multiple stages, with each stage focusing on a single sub-task. More precisely, our pipeline consists of three stages: {\it action candidate labeling}, {\it action naming} and {\it action trimming}. Figure~\ref{fig:annotation_pipeline} presents an illustration of the stages. Our pipeline creates a more reliable work flow for action annotation. For example, it allows us to identify and correct errors from previous stages. We annotated our dataset in-house to ensure the best quality, although the pipeline and tools can be easily scaled to crowd-source settings. We now describe our annotation pipeline in details.

\noindent \textbf{Action Candidate Labeling}. This stage aims to identify all potential actions and their rough temporal extents. We use ELAN~\cite{wittenburg2006elan}\footnote{ELAN is a multi-modal annotation tool developed at Max Planck Institute for Psycholinguistics, The Language Archive, Nijmegen, The Netherlands. \url{http://tla.mpi.nl/tools/tla-tools/elan/}} for this step. Specifically, annotators were ask to mark rough onsets and offsets of possible actions in a video. Two action candidates were allowed to overlap in time. A small to moderate amount of errors are expected in trade of efficiency, as later stages will refine the candidates.

\noindent \textbf{Action Naming}. This stage seeks to label action candidates from previous stage under our taxonomy. We cropped all action candidates into video clips with temporal padding on both the start and the end. These clips are most likely to include the full extent of a single action. Cropping the videos not only reduces the visual content that an user has to examine, but also helps to ensure that the name of the action can be inferred using an isolated clip. We then developed a web interface that presents the clips to the annotators and allows the annotators to input the action names. 

Specifically, the web interface presents one clip at a time and asks the users to annotate a verb and a set of nouns. For the verb, a user can choose from a list 19 pre-defined words. We empirically verified that our list provides a good coverage of actions in our videos. For nouns, a user must enter them in a list. Moreover, we allowed users to red flag a clip if (1) no actions are presented; or (2) there are multiple major actions; or (3) the action is not complete in the given clip. Flagged clips were manually examined and removed if they do not contain a proper action instance.

\noindent \textbf{Action Trimming}. This final stage further refines the temporal boundaries of labeled action clips from the previous stage. Similar to action naming, we developed a web interface where both a video clip (temporally padded) and its action label are presented to annotators. The users were instructed to identify the exact temporal extent of the labeled action. This is done by marking the onset and offset on a slider bar, as shown in Figure~\ref{fig:annotation_pipeline} (right). Similarly, we allowed the users to flag action clips with incorrect labels. Flagged clips were checked manually.

\noindent \textbf{Post-Processing}. We post-processed the labels to finalize the annotations. First, we removed action clips that are less than 0.5 seconds, as (1) the frames were typically blurred due to rapid motion; (2) it was hard to accurately identify their temporal boundaries. Moreover, we ran a spell checker to nouns and merged all the synonyms. In addition, we combined some sub-categories of objects into its super categories to fix naming inconsistency among our annotators. Specifically, fork, knife and spoon were merged into ``eating utensil''. Spatula, skimmer and ladle were renamed as ``cooking utensil'' and jar, bottle and box were renamed ``container''. Finally, action categories were sorted by their frequency. The top 106 categories were considered for our final dataset.

\noindent \textbf{Action Label Statistics}. After post-processing, our action annotation includes $19$ verbs, $53$ nouns and $106$ unique action labels (a combination of verb and nouns), leading to a total of 10,325 action instances within all 86 videos. Figure~\ref{fig:new_data_stats} shows all our verbs, as well as top $10$ objects and top $20$ action labels. While the combination of verb and nouns can be lead to complex actions, the most frequent actions tend to be simple, such as ``read recipe'' or ``open fridge''.

Moreover, we present the distribution of verbs, nouns and action labels in Figure~\ref{fig:new_data_stats}. Figure~\ref{fig:new_data_stats} demonstrates the a ``long tailed'' distribution of our action categories. While the most common action ``read recipe'' happens 752 times, the least common ``put (down) oil container'' only occurs 32 times. This distribution, which we believe have characterized our visual experience, is drastically different from previous action recognition datasets, such as UCF101 or HMDB. This poses a significant challenge of learning from unbalanced samples.

\begin{table*}[t]
\setlength{\tabcolsep}{4pt}
\centering{
\caption{\newedit{Comparison of FPV datasets. Our
EGTEA Gaze+ is the only large scale dataset that offers gaze tracking, hand masks and action annotations at the same time. $^\dagger$: resolution may vary.}}
\begin{tabular}{c|c|c|c|c|c|c|c|c|c|c}
\multirow{2}{*}{FPV Dataset} & \multirow{2}{*}{Mounting} & \multirow{2}{*}{Res (FPS)} & \multirow{2}{*}{Hours} & \multirow{2}{*}{Sessions} & \multirow{2}{*}{Subjects} & \multirow{2}{*}{\begin{tabular}[c]{@{}c@{}}Action\\ Instance\end{tabular}} & \multirow{2}{*}{\begin{tabular}[c]{@{}c@{}}Action\\ Classes\end{tabular}} & \multirow{2}{*}{Hand} & \multirow{2}{*}{Gaze} & \multirow{2}{*}{\begin{tabular}[c]{@{}c@{}}Object\\ Boxes\end{tabular}} \\
         &       &        &    &       &       &                &               &   &   &             \\ \hline \hline
EGTEA Gaze+                  & Head  & 1280*960 (24)              & 28                     & 86    & 32    & 10,321         & 106           & $\checkmark$  & $\checkmark$  & $\times$            \\
EPIC-Kitchens~\cite{Damen2018EPICKITCHENS}                & Head                     & 1920*1080$^\dagger$(60)             & 55                     & 432   & 32    & 39,596         & 149           & $\times$  & $\times$  &  $\checkmark$           \\
Charades-Ego~\cite{sigurdsson2018charadesego}                & Vary  & Vary   & 34.4(+34.4)            & 7860  & 112   & 68,536         & 157           & $\times$  & $\times$  &  $\times$           \\
GTEA Gaze+~\cite{li2015delving}                  & Head  & 1280*960 (24)              & 9  & 37    & 6     & 3,371          & 44            & $\times$  &  $\checkmark$ &  $\times$           \\
UCI ADL~\cite{pirsiavash2012detecting}  & Chest                     & 1280*960 (30)              & 10                     & 20    & 20    & 436            & 32            & $\times$  & $\times$  &  $\checkmark$           \\
JPL Interaction~\cite{ryoo2013first}              & Chest                     & 320*240 (30)               & 1  & 57    & 8     & 94             & 7             & $\times$  & $\times$  & $\times$            \\
CMU MMAC~\cite{Frade-2008-9926}                     & Head  & 640*480 (12)               & 6  & 25    & 5     & 516            & 31            & $\times$  & $\times$  &  $\times$          
\end{tabular}\label{dataset-table}
}\vspace{-1em}
\end{table*}

\subsection{Our EGTEA Gaze+ Dataset}
Our final dataset includes $28$ hours of FPV videos with a resolution of $1280\times 920$ at $24$Hz. The dataset has $86$ unique sessions from $32$ subjects across 7 recipes. Each session consists of a HD video, an audio sampled at $44$KHz, binocular gaze tracking data ($30$Hz), frame-level action annotations and hand masks at sparsely sampled frames. Our annotations include $15$K hand masks and $10321$ action instances from the most frequent $106$ action categories. Our action instances have an average duration of $4.2$ seconds with an average of $11$ events per minutes. That is over $1$ million action frames (from over $2.5$ millions of frames) together with their tracked gaze points. 

\noindent \textbf{Train/Test Splits}. To facilitate fair benchmark using our dataset, we further created three different splits of non-overlapping train and test sets, with 8299/2022, 8299/2022, 8230/2021 samples (train/test). These splits were created by random sampling such that roughly 80\% of the samples per category is used for training and rest for testing. We encourage reporting results on all three splits.

\noindent \textbf{Comparison to FPV Datasets}. Table~\ref{dataset-table} compares our EGTEA Gaze+ dataset with existing FPV action datasets. In comparison to previous FPV datasets~\cite{Frade-2008-9926, ryoo2013first,pirsiavash2012detecting,li2015delving}, our dataset excels at scale. Similar to the concurrent works of EPIC-Kitchens~\cite{Damen2018EPICKITCHENS} and Charades-Ego~\cite{sigurdsson2018charadesego}, our dataset features HD videos and considers similar number of action categories. While EGTEA does have fewer action instances, our dataset stands out in terms of egocentric gaze and hand. \newedit{Notably, our EGTEA is \emph{the only large-scale dataset that offers gaze tracking, hand masks and action annotations at the same time}, thereby providing the most comprehensive benchmark for FPV gaze and actions.} We anticipate that our dataset will be used for FPV gaze estimation, hand segmentation, action recognition and action detection.

% backup
%Our dataset is desgined for FPV action recognition and differs from previous datasets in that: (1) it includes HD video with much higher resolution (960P); (2) it provides action annotations in video sequences and thus facilitates the task of action detection; (3) it offers a diverse set of fine-grained action labels, including 19 unique verbs and 53 unique nouns (objects), leading to a combination of 106 actions; (4) it comes with gaze tracking data for all videos; (5) it elicits the major challenge of long-tailed distribution of actions, which is not available in some of the previous datasets (e.g., UCF101 or HMDB).

%% file: approach.tex
\section{Modeling Gaze and Actions in FPV}
We now present our joint model of egocentric gaze and actions. We denote an input first person video as $x=(x^1, ..., x^t)$ with its frames $x^t$ indexed by time $t$. Our goal is to predict the action category $y$ for $x$. We assume egocentric gaze measurements $g = (g^1, ..., g^t)$ are available during training yet need to be inferred during testing. $g^t$ are measured as a single 2D gaze point at time $t$ defined on the image plane of $x^t$. For our model, it is helpful to reparameterize $g^t$ as a 2D saliency map $g^t(m,n)$, where the value of the gaze position are set to one and all others are zero. And thus $\Sigma_{m,n} g^t(m,n)=1$. In this case, $g^t(m,n)$ defines a proper probabilistic distribution of 2D gaze. 

\begin{figure*}[t]
\centering
\includegraphics[width=0.75\linewidth]{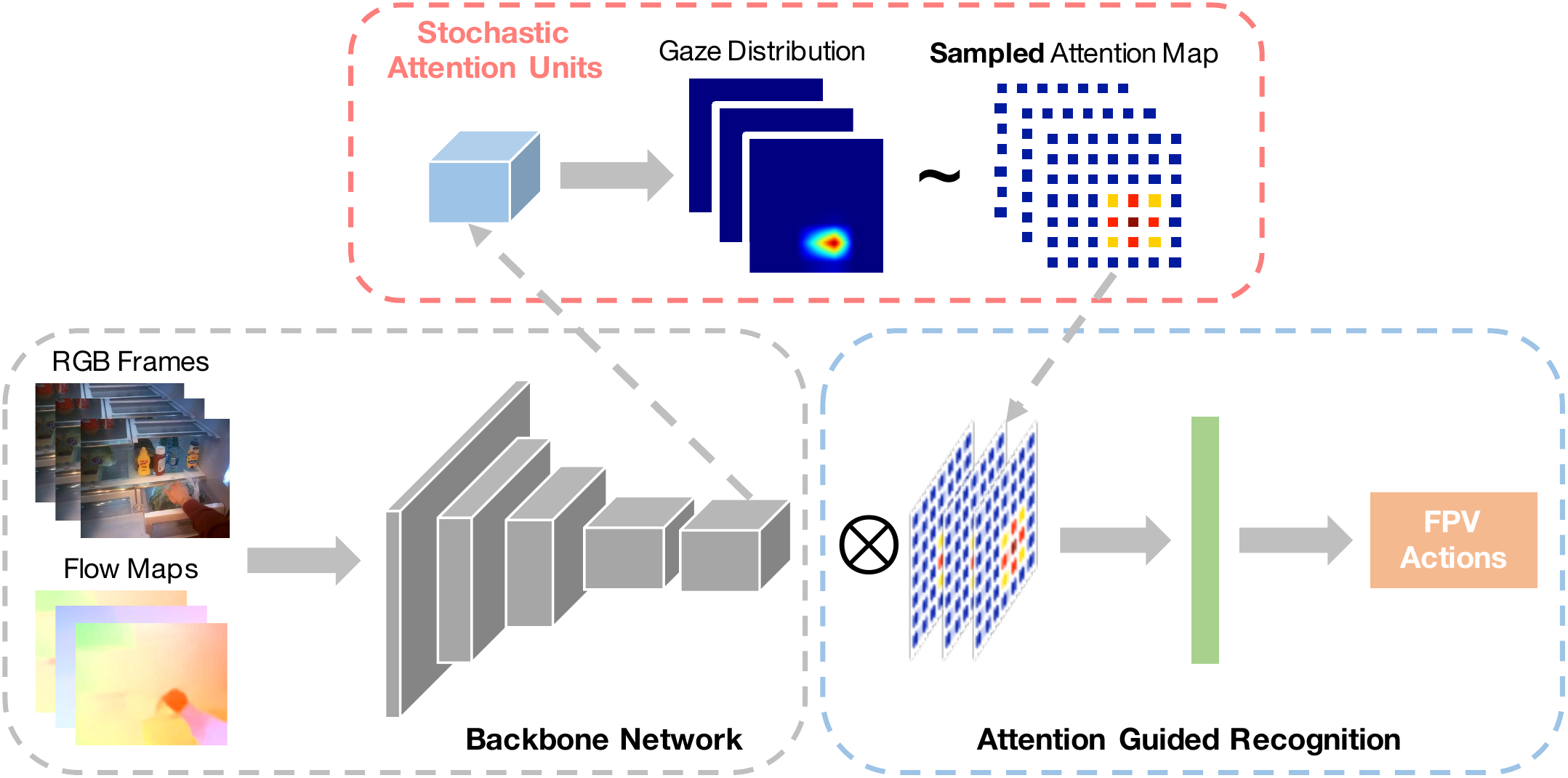}
\caption{Overview of our joint model of FPV gaze and actions. Our model takes multiple RGB and flow frames as inputs, and outputs a set of parameters defining a distribution of gaze in the middle layers. We then sample a gaze map from this distribution. This map is used to selectively pool visual features at higher layers of the network for action recognition. During training, our model receives action labels and noisy gaze measurement. Once trained, the model is able to infer gaze and recognize actions in FPV. We show that this network builds a probabilistic model that naturally accounts for the uncertainty of gaze and captures the relationship between gaze and actions in FPV.} 
\label{fig:overview}\vspace{-1em}
\end{figure*}

Figure~\ref{fig:overview} presents an overview of our model. Consider an analogy between our model and the well-known R-CNN framework for object detection~\cite{girshick2014rich,NIPS2015_5638}. Our model takes a video $x$ as input and outputs the distribution of gaze $q$ as an intermediate result. We then sample the gaze map $g$ from this predicted distribution. $g$ encodes location information for actions and thus can be viewed as a source of action proposals---similar to the object proposals in R-CNN. Finally, we use the attention map to select features from the network hierarchy for recognition. This can be viewed as Region of Interest (ROI) pooling in R-CNN, where visual features in relevant regions are selected for recognition.

\subsection{Modeling Gaze with Stochastic Units}
Our key idea is to model $g(m,n)$ as a probabilistic variable to account for its uncertainty. More precisely, we model the conditional probability of $p(y | x)$  by
\begin{equation}
p(y | x) = \int_{g} p(y | g, x) p(g | x) dg.
\end{equation}
Intuitively, $p(g|x)$ estimates gaze $g$ given the input video $x$. $p(y|g, x)$ further uses the predicted gaze $g$ to select visual features from input video $x$ to predict the action $y$. Moreover, we want to use high capacity models, such as deep networks, for both $p(g|x)$ and $p(y|g,x)$. While this model is appealing, the learning and inference tasks are intractable for high dimensional video inputs $x$.

Our solution, inspired by~\cite{kingma2013auto,sohn2015learning}, is to approximate the intractable posterior $p(g | x)$ with a carefully designed $q_{\pi}(g| x)$. Specifically, we define $q(m,n) \in R$ defined discrete $2$D grids, e.g., the image plane. $q(m,n)$ has the same size of $M\times N$ as $x$. $q$ is parameterized by $\pi_{m, n}$, where
\begin{equation}
q(m, n) = q( g_{m, n} = 1 | x ) = \frac{\pi_{m,n}}{\sum_{m,n} \pi_{m, n}}.
\end{equation}
$\pi = q_\psi(x)$ is the output from a deep neural network $q_\psi$. $q( g | x)$ thus models the probabilistic distribution of egocentric gaze. Thus, our deep network creates a 2D map of $\pi_{m,n}$. $\pi$ defines an approximation $q_\pi$ to the distribution of the latent attention map. Specifically, $q(m, n)$ can be viewed as the expectation of the gaze $g$ at position $(m, n)$. We can then sample the gaze map $\tilde{g}$ from $q_\pi$ for recognition. 

Given a sampled gaze map $\tilde{g}$, our attention mechanism will selectively aggregate visual features $\phi(x)$ defined by network $\phi$. In our model, this is simply a weighted average pooling, where the weights are defined by the gaze map $\tilde{g}$. We then send pooled features to the recognition network $f$. We further constrain $f$ to have the form of a linear classifier, followed by a softmax function. This design is important for approximate inference. Now we have
\begin{equation}
\begin{split}
   p(y | g, x) &= f(\Sigma_{m,n} \tilde{g}_{m,n} \phi(x)_{m,n}) \\
               &= softmax\left(W_f^T (\Sigma_{m,n} \tilde{g}_{m,n} \phi(x)_{m,n})\right).
\end{split}
\end{equation}
The sum operation is equivalent to spatially re-weighting individual feature channels. We expect that the network will learn to attend to discriminative regions for action recognition. Note that this is a soft attention mechanism that allows back-propagation. Thus, top-down modulation of gaze can be achieved through gradients from action labels.

Our model thus includes three sub-networks: $q_\psi(x)$ that outputs parameters for the attention map, $\phi(x)$ that extracts visual representations for $x$, and $f(g, x)$ that pools features and recognizes actions. All three sub-networks share the same backbone network with their separate heads, and thus our model is realized as a single feed forward deep network. Due to the sampling process introduced in modeling, learning the parameters of the network is challenging. We overcome this challenge by using variational learning and optimizing a lower bound. We now present our training objective and inference method. 

\subsection{Variational Learning}
During training, we make use of the input video $x$, its action label $y$ and human gaze measurements $g$ sampled from a distribution $p(g|x)$. Intuitively, our learning process has two major goals. First, our predicted gaze distribution parameterized by $q_\psi(x)$ should match the noisy observations of gaze. Second, the action recognition error should be minimized. We achieve these goals by maximizing the lower bound of $\log p(y|x)$, given by
\begin{equation}
\begin{split}
    \log p(y|x) \geq -\mathcal{L} & = E_{g\sim q(g|x)}[\log p(y|g,x)] \\
                                  & - KL[q(g|x)||p(g|x)],
\end{split}
\end{equation}
where $KL(p||q)$ is the Kullback--Leibler (KL) divergence between $p$ and $q$, and $E$ denotes the expectation.

\noindent\textbf{Learning with Egocentric Gaze}. Computing $KL(p||q)$ requires the prior knowledge of $p(g|x)$. In our case, given $x$, we observe gaze $g$ drawn from $p(g|x)$. Thus, $p(g|x)$ is the noise pattern of the gaze measurement $g$. We adapt a simple noise model of gaze. For all tracked fixation points, we assume a $2$D isotropic Gaussian noise, where the standard deviation of the Gaussian is selected based on the average tracking error of modern eye trackers. When the gaze point is a saccade (or is missing), we set $p(g|x)$ to a $2$D uniform distribution, allowing attention to be uniformly distributed. 

\noindent\textbf{Learning without Human Gaze}.
\newedit{Modeling $p(g|x)$ using human gaze requires gaze tracking data for training. This is not feasible for datasets without gaze, such as EPIC-Kitchens~\cite{Damen2018EPICKITCHENS}. As an extension to our previous work~\cite{li2018eccv}, we propose to model $p(g|x)$ using a simple prior $p(g)$ --- a $2$D uniform distribution. This prior assumes that the attention can be allocated with equal chance to every location on the image plane. We demonstrate that even such a weak prior helps to regularize the learning when gaze data is not available. Other distributions can be further explored.}

\noindent\textbf{Loss Function}. Given our prior of gaze $p(g|x)$, we now minimize our loss function as the negative of the empirical lower bound, given by
\begin{equation}
\textstyle
-\sum_g \log p(y|g, x) + KL[q(g|x)||p(g|x)].
\end{equation}
During training, we sample the gaze map $\tilde{g}$ from the predicted distribution $q(g|x)$, apply the map for recognition ($p(y|\tilde{g}, x)=f(\tilde{g}, x)$) and compute its negative log likelihood---the same as the cross entropy loss for a categorical variable $y$. Our objective function thus has two terms: (a) the negative log likelihood term as the cross entropy loss between the predicated and the ground-truth action labels using the sampled gaze maps; and (b) the KL divergence between the predicted distribution $q(g|x)$ and the gaze distribution $p(g|x)$. 

\noindent\textbf{Reparameterization}. Our model is fully differentiable except for the sampling of $\tilde{g}$. To allow end-to-end back propagation, we re-parameterize the discrete distribution $q(m, n)$ using the Gumbel-Softmax approach as in~\cite{jang2016categorical,maddison2016concrete}. Specifically, instead of sampling from $q(m, n)$ directly, we sample the gaze map $\tilde{g}$ via
\begin{equation}
\tilde{g}_{m, n} \sim \frac{\exp ((\log \pi_{m,n} + G_{m,n})/\tau)}{\sum_{m, n} \exp ((\log \pi_{m,n} + G_{m,n})/\tau)},
\end{equation}
where $\tau$ is the temperature that controls the ``sharpness'' of the distribution. We set $\tau=2$ for all of our experiments. The softmax normalization ensures that $\sum_{m,n} \tilde{g}(m, n) = 1$, such that it is a proper gaze map. $G$ follows the Gumbel distribution $G = -\log(-\log{U})$, where $U$ is the uniform distribution on $[0, 1)$. This Concrete distribution separates out the sampling into a random variable from a uniform distribution and a set of parameters $\pi$, and thus allows the direct back-propagation of gradients to $\pi$.

\subsection{Approximate Inference}
During testing, we feed an input video $x$ forward through the network to estimate the gaze distribution $q(g|x)$. Ideally, we should sample multiple gaze maps $\tilde{g}$ from $q$, pass them into our recognition network $f(g,x)$, and average all predictions. This is, however, prohibitively expensive. Since $f(g, x)$ is nonlinear and $g$ has hundreds of dimensions, we will need many samples $\tilde{g}$ to approximate the expectation $E_g[f(g, x)]$, where each sample requires us to recompute $f(\tilde{g}, x)$. We take a shortcut by feeding $q_\pi$ into $f$ to avoid the sampling. We note that $q_\pi$ is the expectation of $\tilde{g}$, and thus our approximation is $E_g[f(g, x)] \approx f(E[g], x)$.

This shortcut does provide a good approximation. Recall that our recognition network $f$ is a softmax linear classifier. Thus, $f$ is convex (even with the weight decay on $W_f$). By Jensen's Inequality, we have $E_g[f(g, x)] \ge f(E[g], x)$. Thus, our approximation $f(E[g], x)$ is indeed a lower bound for the sample averaged estimate of  $E_g[f(g, x)]$. Using this deterministic approximation during testing also eliminates the randomness in the results due to sampling. We have empirically verified the effectiveness of our approximation.

\subsection{Discussions}
We connect our model to the techniques of Dropout and DropBlock, as well as the model of Conditional Variational AutoEncoder (CVAE). We hope these connections help to draw better insights about our model. 

\noindent\textbf{Connection to Dropout and DropBlock}. Our sampling procedure during learning can viewed as an alternative to Dropout and the more recent DropBlock~\cite{srivastava2014dropout,ghiasi2018dropblock}, and thus helps to regularize the learning. In particular, we sample the gaze map $\tilde{g}$ to re-weight features. This map will have a single peak and many close-to-zero values because of the softmax function. If a position $(m,n)$ has a very small weight, the features at that position are effectively ``dropped out''. The key difference is that our sampling is guided by the predicted gaze distribution of $q_\psi$ instead of random masking used by Dropout or DropBlock. 

\noindent\textbf{Connection to Conditional Variational Autoencoder}. Our model is highly relevant to CVAE~\cite{sohn2015learning}. Both models use stochastic variables for discriminative tasks. Yet they are different in three aspects: (1) our stochastic unit---the 2D gaze distribution, is discrete. In contrast, CVAE employs a continuous Gaussian variable, leading to a different reparameterization technique. (2) our stochastic unit---the gaze map is physically meaningful and \emph{receives supervision} during training, while CVAE's is latent. (3) our model approximates the posterior with $q_\psi(x)$ and uses one forward pass for approximated inference, while CVAE models the posterior as a function of both $x$ and $y$ and requires recurrent updates. 

\subsection{Network Architecture}
Our model builds on two-stream I3D networks~\cite{carreira2017quo}. Similar to its base Inception network~\cite{szegedy2015going}, I3D has 5 convolutional blocks and the network uses 3D convolutions to capture the temporal dynamics of videos. Specifically, our model takes both RGB frames and optical flow as inputs, and feeds them into an RGB or a flow stream, respectively. We fuse the two streams at the end of the 4th convolutional block for gaze estimation, and at the end of the 5th convolutional block for action recognition. The fusion is done using element-wise summation as suggested by~\cite{feichtenhofer2016convolutional}. We used 3D max pooling to match the predicted gaze map to the size of the feature map at the 5th convolutional block for weighted pooling. 

Our model takes the inputs of 24 frames, outputs action scores and a gaze map at a temporal stride of 8. Our output gaze map has a spatial resolution of $7\times 7$ (downsampled by 32x). During testing, we average the clip-level actions scores to recognizing actions in a video. Note that our gaze output is down-sampled both spatially (x32) and temporally (x8). When evaluating gaze, we aggregate fixation points within 8 frames and project them into a downsampled 2D map. This time interval (300ms) is equal to the duration of a fixation (around 250ms) and thus this temporal aggregation should preserve the location of gaze.

\noindent\textbf{Implementation Details}. We downsampled all video frames to $320\times 256$ and computed optical flow using FlowNet V2~\cite{ilg2016flownet}. The flow map was truncated in the range of $[-20, 20]$ and rescaled to $[0, 255]$ as~\cite{wang2016temporal,simonyan2014two}. During training, we randomly cropped $224\times 224$ regions from $24$ frames. We then fed the RGB frames and flow maps into our networks. We also performed random horizontal flip and color jittering for data augmentation. For testing, we send the frames with a resolution of $320\times 256$ and their flipped version. For action recognition, we averaged pool scores of all clips within a video. For gaze estimation, we flipped back the gaze map and take the average.

\noindent\textbf{Training Details}. All our models are trained using SGD with momentum of 0.9 and weight decay of $0.00004$. When using SGD for variational learning, we draw a single sample for each input within a mini-batch, and multiple samples of the same input will be drawn at different iterations. The initial weights for 3D convolutional networks are restored from Kinectcs pre-trained models~\cite{carreira2017quo}. For training two stream-networks, we used a batch size of 40, paralleled over 4 GPUs. We used a initial learning rate of $0.032$, which matches the same learning rate from~\cite{carreira2017quo}. We decayed the learning rate by a factor of $10$ at $40$th epoch and end the training at $80$ epochs. We enabled batch normalization~\cite{ioffe2009batch} during training and set the decay rate for its parameters to $0.9$, allowing faster aggregation of dataset statistics. \newedit{Different from our previous work~\cite{li2018eccv}, dropout with rate of $0.7$ was attached for the fully connected layer during training. We found that adding dropout and training for longer help to prevent overfitting and thus lead to improved results.}

%% file: result.tex
\section{Experiments and Results}
\label{sec:result}

This section presents our experiments and results. We first introduce the datasets and the evaluation metrics. We then present our results on gaze and actions using EGTEA Gaze+ and EPIC Kitchens datasets. Specifically, our main results include three parts. First, we present an ablation study of our model. Second, we present result on gaze estimation and compare to a set of strong baselines on EGTEA Gaze+ dataset, where the ground truth gaze points are available. Finally, we discuss our main results on FPV action recognition. Our results are compared to several state-of-the-art methods on both EGTEA Gaze+ and EPIC-Kitchens datasets. Overall, our model achieves strong results for both gaze estimation and action recognition. 

\subsection{Dataset and Evaluation Metric}
We start by introducing the datasets used in our experiment, and present the evaluation metrics used on these datasets.

\noindent \textbf{Dataset}. We use EGTEA Gaze+ dataset as the main vehicle for our benchmark. The dataset includes $10,321$ action instances from $106$ categories. These instances are divided into three train and test splits. \newedit{For our experiments, we report gaze estimation and action recognition results on all splits}. EGTEA Gaze+ manifests two key challenges of FPV action recognition. First, the definition of FPV actions leads to a {\it fine-grained recognition problem}. The task is to recognize action categories like ``cut onion'' or ``spread condiment (on) bread (using) eating utensil''. Thus, the inference of these categories involves non-trivial understanding of body motion and object information. Moreover, the action labels are imbalanced --- the distributions of action instances in both datasets follow a {\it long-tailed distribution}. The frequent classes have hundreds of samples and the classes on the tail have only dozens of samples. This long-tailed distribution poses new challenge of learning from imbalanced data.

\noindent\textbf{Evaluation Metric}. We now present our evaluation metrics for FPV gaze estimation and action recognition.

\noindent {\it Gaze Estimation}: We consider gaze estimation as a binary classification problem, evaluate their Precision and Recall curves, and report the best F1 scores together with their corresponding precision and recall. Untracked gaze points and saccades are excluded from the evaluation. Note that we did not use the average angular error common in gaze tracking. This is because that our model produces a low resolution gaze map. The gaze estimation results are only reported on EGTEA Gaze+ where the ground truth gaze is given by a mobile eye tracker.

\begin{table}[t]
\centering
\caption{Ablation study of backbone networks on EGTEA Gaze+ dataset. We compare RGB, Flow, late fusion and joint training of I3D for action recognition, and report mean class accuracy \newedit{across all splits}.}\label{tab:ablation:network}
\begin{tabular}{c|ccc}
\multirow{2}{*}{} 
\newedit{Method} & 
\multicolumn{3}{c}{\newedit{Mean Class Accuracy}} \\
\multicolumn{1}{c|}{}   & \newedit{Split1} & \newedit{Split2} &\newedit{Split3}          \\ \hline 
\newedit{I3D RGB} & \newedit{50.08} &\newedit{49.01} &\newedit{46.89}   \\
\newedit{I3D Flow} &\newedit{45.78} &\newedit{42.68} &\newedit{41.36}   \\
\newedit{I3D Fusion} &\newedit{54.19} &\newedit{51.45} &\newedit{49.41}   \\
\newedit{I3D Joint}  &\newedit{\textbf{55.76}} &\newedit{\textbf{53.14}} &\newedit{\textbf{53.55}} 
\end{tabular}\vspace{-1em}
\end{table}

\noindent {\it Action Recognition}: We treat action recognition as a multi-class classification problem. For EGTEA Gaze+, we report mean class accuracy --- the average per-class accuracy for action instances across all three splits.
Mean class accuracy provides a fair metric in comparison to instance level accuracy for class-imbalanced problems. \newedit{Mean class accuracy should not be confused with instance-level accuracy. The two metrics are different and not directly comparable.}

\subsection{Ablation Study}
We begin with a comprehensive study of our model on the first split of the EGTEA Gaze+ dataset. The goal of this ablation study is to delineate different components of our model. Specifically, we vary each of the following components: (1) the backbone network for feature presentation; (2) probabilistic modeling; and (3) the attention guided action recognition, and evaluate their impact to the performance.

\begin{table*}[t]\centering
\caption{Ablation study of probabilistic modeling on EGTEA. \newedit{We compare variants of our probabilistic model to its deterministic version (Gaze MLE) across all three splits. We report F1 scores for gaze and mean class accuracy for action. N/A indicates that model does not output gaze. }}\label{tab:ablation:prob}
\begin{tabular}{c|cc|cc|cc}
\multicolumn{1}{c|}{\multirow{2}{*}{Method}}          
&\multicolumn{2}{c|}{Split1} &\multicolumn{2}{c|}{\newedit{Split2}} &\multicolumn{2}{c}{\newedit{Split3}} \\ \cline{2-7}
\multicolumn{1}{c|}{}   & Gaze F1 & Action Acc & \newedit{Gaze F1} & \newedit{Action Acc} & \newedit{Gaze F1} & \newedit{Action Acc}\\ \hline 
I3D Joint & N/A &55.76 &\newedit{N/A} &\newedit{53.14} &\newedit{N/A} & \newedit{53.55}   \\
%Prob-Center &N/A &53.10 &N/A &50.12 &N/A &49.98  \\
Gaze MLE & 26.63 &55.88 &\newedit{24.65} & \newedit{52.77} & \newedit{29.57} & \newedit{53.49}   \\
Ours (Prob. w. Gaze) & {\bf 34.01}  &{\bf 57.20} &\newedit{{\bf 31.79}} &\newedit{{\bf 53.75}} &\newedit{{\bf 35.73}} &\newedit{{\bf 54.13}} \\
\newedit{Ours (Prob. wo. Gaze)} &\newedit{11.63} &\newedit{56.50} &\newedit{10.17} &\newedit{53.52} &\newedit{12.98} &\newedit{53.58} 
\end{tabular}\vspace{-1em}
\end{table*}

\begin{table*}[t]\centering
% main caption
\caption{\textbf{Gaze estimation results on EGTEA}. We report best F1 scores and their corresponding precision and recall. We compare our model to several baselines. Our results are comparable to the latest methods. $^\dagger$:  methods jointly model gaze and actions; *: see Section~\ref{sec:gaze_results} for discussions.}\label{tab:results:gaze}
\begin{tabular}{c|ccc|ccc|ccc}
\multicolumn{1}{c|}{\multirow{2}{*}{Method}}          
&\multicolumn{3}{c|}{Split1} &\multicolumn{3}{c|}{Split2} &\multicolumn{3}{c}{Split3} \\ \cline{2-10}
\multicolumn{1}{c|}{}   & Prec & Recall  & F1   &Prec  &Recall & F1 & Prec &Recall &F1 \\ \hline 
\newedit{Center Prior}   & \newedit{6.39} & \newedit{31.96} & \newedit{10.65} &\newedit{4.55} &\newedit{22.77} &\newedit{7.59} & \newedit{7.16} &\newedit{35.82} &\newedit{11.94}\\
EgoGaze*~\cite{li2013learning}    & 16.63 & 16.63 & 16.63 &12.85 &12.85 &12.85 & 18.30 &18.30 &18.30\\
Simple Gaze         & 16.11  & 41.82  & 31.33 &24.66	&37.16	&29.65 &29.52	&38.85	&33.55\\											               
Deep Gaze~\cite{zhang2017deep}     &{\bf28.71}	&{\bf43.08}	&{\bf 34.46} &{\bf26.30} &40.28	&{\bf  31.82} &{\bf30.96}	&{\bf 44.48}	&{\bf 36.51}\\ 
Gaze MLE$^\dagger$           &21.25	&35.65	&26.63 &18.15	&38.40	&24.65 &24.11	&38.23	&29.57\\ 
Ours$^\dagger$          &28.29	&42.65	&34.01 &26.05	&{\bf40.76}	&31.79 &30.89	&42.37	&35.73
\end{tabular}\vspace{-1em}
\end{table*}

\noindent \textbf{Backbone Network: RGB vs.\ Flow}. We first benchmark different backbone networks for FPV action recognition. Concretely, we tested RGB and flow streams of I3D~\cite{carreira2017quo}, the late fusion of two streams, and the joint training of two streams~\cite{feichtenhofer2016convolutional}. The results are summarized in Table~\ref{tab:ablation:network}. Overall, EGTEA dataset is challenging, even the strongest model has an accuracy around $55\%$. To help calibrate the performance, we note that the same I3D model achieved $36\%$ on Charades~\cite{sigurdsson2016hollywood}, $74\%$ on Kinetics and $99\%$ on UCF~\cite{carreira2017quo}. 

Unlike Kinetics or UCF, where the flow stream performs comparably to the RGB stream, \newedit{the performance of I3D flow on EGTEA is significantly worse (4.3\%-6.3\%) than I3D RGB. The results suggest that motion cues are more difficult to capture in FPV. This is probably due to the frequent camera motion in FPV. The joint training of RGB and flow streams performs the best across all splits.} We thus choose this network as our backbone for the rest of our experiments.

\noindent\textbf{Modeling: Probabilistic vs.\ Deterministic}. Going forward, we test the probabilistic modeling part of our method. We focus on the key question: ``Does probabilistic modeling of gaze helps?'' To this end, we compare to a deterministic version of our model that uses maximum likelihood estimation for gaze. We denote this model as \emph{Gaze MLE}. Instead of sampling, this model learns to directly output a gaze map, and applies the map for recognition. During training, the gaze map is supervised by human gaze using a pixel-wise sigmoid cross entropy loss. Both model architecture and the training procedure are the same as our model. We disable the loss for gaze when fixation is not available. 

We compare our model with Gaze MLE for gaze and actions, and present the results in Table~\ref{tab:ablation:prob}. \newedit{Gaze MLE is on par with the backbone I3D joint and worse than our model. Across all splits, our probabilistic model outperforms its deterministic version by an average of $1.0\%$ for action recognition and $6.9\%$ for gaze estimation. We conjecture that our probabilistic model helps to facilitate the learning even with a noisy supervisory signal such as human gaze.} 

\newedit{\noindent\textbf{Training without Human Gaze}. Finally, we report the results of our model using the same backbone yet trained without using human gaze in Table~\ref{tab:ablation:prob}). This version follows the same network architecture yet receives uniform distribution as the prior for variational learning. For action recognition, this model is slightly worse than gaze supervised models by an average of $0.5\%$, yet marginally outperforms the base I3D and the Gaze MLE by $0.4\%$ and $0.5\%$ respectively. For gaze estimation, this model lags far behind of the gaze supervised model (-$23\%$). These results suggest that (1) probabilistic attention helps to improve action recognition even without explicit supervision of gaze; and (2) adding human gaze as supervision provides a moderate performance gain for actions and a significant gain for gaze.}

\newedit{\noindent\textbf{Improvements to Our Conference Paper}. In comparison to our conference paper~\cite{li2018eccv}, adding dropout and training for longer improve both the baseline I3D and our model. For action recognition, the mean class accuracy of I3D joint increases from $49.79$ to $53.55$ on the first split. Similarly, the accuracy of our model increases from $53.30$ to $57.20$. For gaze estimation, the F1 score of our model also improves from $32.66$ to $36.01$. Importantly, we demonstrate that our model trained without human gaze can still outperform I3D.}

\subsection{FPV Gaze Estimation}\label{sec:gaze_results}
We now present our results for FPV gaze estimation.

\noindent\textbf{Baselines}. We consider the following baselines.
\begin{itemize}
    \item \newedit{{\bf Center Prior} uses a fixed attention map by a 2D Gaussian distribution with its parameters estimated on the training gaze data. This is a simple baseline.}  
    \item {\bf EgoGaze~\cite{li2013learning}} makes use of hand crafted egocentric features, such as head motion and hand position, to regress gaze points. For a fair comparison, we use the FlowNet V2 for motion estimation and hand masks from FCN for hand positions (same as our method). EgoGaze outputs a single gaze point per frame. With a single ground-truth gaze, EgoGaze will have equal numbers of false positives and false negative. Thus, its precision, recall and F1 scores are the same.
    \item {\bf Simple Gaze} is a deep model for gaze estimation. Specifically, we directly estimate the gaze map using per-pixel sigmoid cross entropy loss. We use the same backbone network (I3D Joint) as our model and keep the output resolution the same.
    \item {\bf Deep Gaze~\cite{zhang2017deep}} is the FPV gaze prediction module from~\cite{zhang2017deep}, where a 3D convolutional network is combined with a KL loss. Again, we use I3D Joint as the backbone network and keep the output resolution. Note that this model can be considered as a special case of our model by removing the sampling, the attention mechanism and the recognition network.
    \item {\bf Gaze MLE} is the same model in our ablation study, where gaze is estimated using maximum likelihood.
\end{itemize}

\noindent\textbf{Results}. Our gaze estimation results are shown in Table~\ref{tab:results:gaze}. We report best F1 scores and their corresponding precision and recall. Not surprisingly, deep models outperform \newedit{the center prior as well as models using hand crafted features} by a large margin. We also observe that models with KL loss (e.g., Deep Gaze and our model) are consistently better than those use cross entropy loss (e.g., Simple Gaze and Gaze MLE). We conjecture that this is due to the difficulty in balancing between the losses for gaze and action. \newedit{Finally, our method has slightly worse results than Deep Gaze (-$0.4\%$ in F1 across three splits). However, our model is able to jointly estimate gaze and recognize actions, while Deep Gaze can only output gaze distribution.}  

\begin{figure*}[t]
\centering
\includegraphics[width=0.9\linewidth]{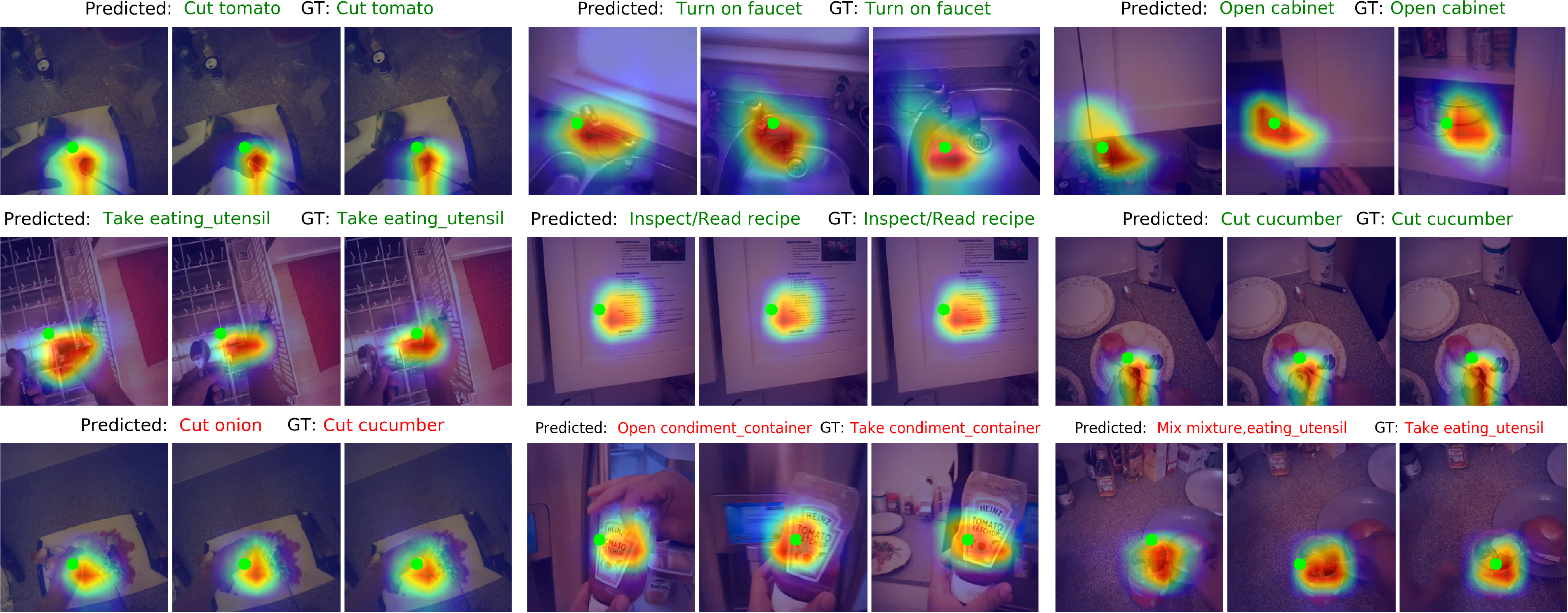}
\caption{Visualization of gaze estimation and action recognition results. For each 24-frame video snippet, we plot the output gaze heat map (higher values in red) with a temporal stride of 8 frames. We also display the ground-truth gaze points as green dots. Thus, the result for each snippet is shown as three key frames with their gaze maps. We print the predicted action labels and ground-truth labels above the images. Both successful (first and second rows) and failure cases (third row) are presented.} 
\label{fig:gaze_action} \vspace{-1em}
\end{figure*}

\noindent\textbf{Discussion}. Our results suggest that the top-down, task-relevant attention is not fully captured in our joint model, even though the top-down modulation can be achieved via back-propagation. This is thus an interesting future direction for the community to explore. Finally, we note that the benchmark of gaze estimation uses noisy human gaze as ground truth. We argue that even though these gaze measurements are noisy, they largely correlate with the underlie signal of attention. And thus the results of the benchmark are meaningful.

\subsection{FPV Action Recognition}
We now describe our results on FPV action recognition. We introduce our baselines, compare our model to the baselines, and discuss the results.

\begin{table}[t]
\footnotesize 
\caption{\textbf{Action recognition results on EGTEA}. \newedit{Results are reported using \emph{mean class accuracy} across all three splits. Our method outperforms most of previous methods except~\cite{kapidis2019multitask}, which uses additional hand masks during training. $^\dagger$: methods use human gaze during testing. -: code not available to produce the results.} }\label{tab:results:action}
\centering
{
\tablestyle{2pt}{1.0}
\setlength{\tabcolsep}{8pt} 
\renewcommand{\arraystretch}{1} % Default value: 1
\begin{tabular}{c|cccc}
\multirow{2}{*}{} 
\newedit{Method} & 
\multicolumn{4}{c}{\newedit{Mean Class Accuracy}} \\
\multicolumn{1}{c|}{}   & \newedit{Split1} & \newedit{Split2} & \newedit{Split3} & \newedit{Avg}         \\ \hline 
\newedit{EgoIDT+Gaze$^\dagger$~\cite{li2015delving}}   & \newedit{42.55} & \newedit{37.30} & \newedit{37.60} &\newedit{39.13}\\
\newedit{2SCNN~\cite{simonyan14very}} & \newedit{43.78} & \newedit{41.47} & \newedit{40.28} & \newedit{41.84} \\
\newedit{I3D (joint)~\cite{carreira2017quo}} & \newedit{55.76} & \newedit{53.14} & \newedit{53.55} & \newedit{54.15}\\
\newedit{I3D+Gaze$^\dagger$}  & \newedit{53.74} & \newedit{50.30} & \newedit{49.63} & \newedit{51.22}\\
\newedit{I3D+Center Prior}  & \newedit{52.48} & \newedit{49.44} & \newedit{49.31} & \newedit{50.41}\\
\newedit{I3D+EgoConv~\cite{singh2016first}} & \newedit{54.19} & \newedit{51.45} & \newedit{49.41} & \newedit{51.68}\\ 
\newedit{Ego-RNN-2S~\cite{sudhakaran2018attention}}   & \newedit{52.40} & \newedit{50.09} & \newedit{49.11} & \newedit{50.53}\\
\newedit{LSTA-2S~\cite{sudhakaran2019lsta}}   & \newedit{53.00} & \newedit{-} & \newedit{-} & \newedit{-}\\
\newedit{Mutual Context-2S~\cite{huang2019mutual}} & \newedit{55.70} & \newedit{-} & \newedit{-} & \newedit{-}\\ \hline
%\newedit{Gaze MLE}    & \newedit{55.88} & \newedit{52.77} & \newedit{53.49} & \newedit{54.05}\\ \hline
\newedit{Ours (wo. gaze)}    & \newedit{56.50} & \newedit{53.52} & \newedit{53.58} & \newedit{54.53} \\
\newedit{Ours (w. gaze)}    & \newedit{\textbf{57.20}} & \newedit{\textbf{53.75}} & \newedit{\textbf{54.13}} & \newedit{\textbf{55.03}}\\\hline
\newedit{Multitask~\cite{kapidis2019multitask}} & \newedit{61.40} & \newedit{-} & \newedit{-} & \newedit{57.60}
\end{tabular}}\vspace{-1em}
\end{table}

\noindent\textbf{Baselines}. We consider the following set of baselines for FPV action recognition.
\begin{itemize}
     \item {\bf EgoIDT+Gaze~\cite{li2015delving}} combines egocentric features with dense trajectory descriptors~\cite{wang2011action}. These features are further selected by gaze points, and encoded using Fisher vectors~\cite{perronnin2010improving} for action recognition.
    \item {\bf 2SCNN~\cite{simonyan2014two}} is the two stream networks with VGG16 backbone developed for generic action recognition. 
    \item {\bf I3D (joint)~\cite{carreira2017quo}} is the two stream I3D with joint training. It is a strong baseline for many datasets.
    \item {\bf I3D+Gaze} is inspired by~\cite{li2015delving,fathi2012learning}, where the ground truth human gaze is used to pool features from the last convolutional outputs of the network. For this method, we use the same I3D joint backbone and the same attention mechanism as our model, yet use human gaze for pooling features. When human gaze is not available, we fall back to average pooling.
    \item \newedit{{\bf I3D+Center Prior} is similar to I3D+Gaze, where a fixed 2D Gaussian distribution (center prior) is used as the attention map for pooling network features.}
    \item {\bf I3D+EgoConv~\cite{singh2016first}} adds a stream of egocentric cues for FPV action recognition. This egocentric stream encodes head motion and hand masks, and its outputs are fused with RGB and flow streams. We use Fully Convolutional Network (FCN)~\cite{long2015fully} for hand segmentation, and late fuse the score of egocentric stream with I3D for a fair comparison. This model is trained from scratch.
    \item {\bf Ego-RNN-2S~\cite{sudhakaran2018attention} and LSTA-2S~\cite{sudhakaran2019lsta}} are recent recurrent networks using soft-attention for FPV action recognition. Both models use two-stream networks. \newedit{We used the authors' code to report results of Ego-RNN-2S. Unfortunately, the training code of LSTA-2S was not released. We can only report their updated results on the first split.}\footnote{Available at \url{https://eyewear-computing.org/EPIC_ICCV19/program/ICCV-EPIC-LSTA.pdf}}
    \item \newedit{\textbf{Mutual Context-2S}~\cite{huang2019mutual} presents a joint model of gaze and actions in FPV using two-stream networks. The model alternates between the estimation of gaze and recognition of actions.}
    \item \newedit{\textbf{Multitask}~\cite{kapidis2019multitask} makes use of a multi-stream network that learns to output gaze data, hand masks, objects (noun) and motion (verb) for FPV action recognition.}
%    \item {\bf Gaze MLE} is the deterministic version of our model that has been used in our ablation and gaze results.
\end{itemize}

We were unable to compare against relevant methods from~\cite{ma2016going,fathi2012learning}. These methods require additional object annotations for training, which is not presented in EGTEA dataset. And we must emphasis that our method does not need object or hand information for training or testing.

\noindent\textbf{Results and Discussion}. We report results on all three splits on EGTEA Gaze+ in Table~\ref{tab:results:action}. As expected, all deep models outperform hand crafted features (EgoIDT+Gaze~\cite{li2015delving}) by a large margin. Among the deep models, 3D networks (I3D) outperforms the 2D version. Surprisingly, I3D~\cite{carreira2017quo}, original designed for generic action recognition, provides a strong baseline that outperforms all previous deep models, including the recent methods~\cite{sudhakaran2018attention, sudhakaran2019lsta}. \newedit{However, naively combining egocentric features (EgoConv+I3D), using egocentric gaze directly (I3D+Gaze) or using the center prior as the attention map (I3D+Center Prior) does not improve the results, and indeed decreases the accuracy by $1$-$3.7\%$. These results suggest that proper integration of egocentric cues, such as gaze, is critical for FPV action recognition.}

%There might be several reasons for this performance drop. First, EgoConv~\cite{singh2016first} was designed to capture actions defined by ``gross body motion'', such as ``take'' vs.\ ``put'', while our setting requires fine grained recognition of actions, e.g., ``take cup'' vs.\ ``take plate''. Thus, EgoConv features is less useful. Second, as we argued in the introduction, human gaze can be quite noisy, with over $25\%$ of gaze points irrelevant to actions. Thus, using gaze data directly might hurt the performance.

\newedit{Moreover, our model supervised by human gaze measurements further improves the strong baseline of I3D by an average of +$0.9\%$ across three splits. Even without human gaze, our model slightly outperforms I3D by +$0.4\%$. Importantly, our results consistently outperform several latest methods for FPV action recognition, including Ego-RNN-2S~\cite{sudhakaran2018attention}, LSTA-2S~\cite{sudhakaran2019lsta}. For example, our model outperforms LSTA-2S~\cite{sudhakaran2019lsta} by $4.2\%$ on the first split, and beats Ego-RNN-2S~\cite{sudhakaran2018attention} by an average of $4.5\%$ across three splits. Importantly, none of these model can output gaze distribution. We argue that these results provide a strong support to the stochastic modeling of gaze measurements. Without accounting for this uncertainty, the model is likely distracted by irrelevant regions from the misleading gaze points.}

\newedit{Finally, we compare our model to latest methods that jointly model gaze and actions, including Mutual Context-2S~\cite{huang2019mutual} and Multitask~\cite{kapidis2019multitask}. Our results outperform Mutual Context-2S~\cite{huang2019mutual} by $1.5\%$ on the first split, yet are worse than Multitask~\cite{kapidis2019multitask} by an average of $2.6\%$ across three splits. All three models use gaze data for training. Multitask~\cite{kapidis2019multitask} additionally employs a stronger backbone~\cite{chen2018multi}, adopts hand masks for training, and considers the joint classification of verbs and nouns using a multi-stream network. We postulate that our model might also benefit from these improvements. It will be particularly interesting to explore attention defined by hands. As pointed by our previous work~\cite{li2015delving}, the position of egocentric hands informs the location of hand-object interactions. We hypothesize that integrating hand attention can improve the performance of our model. While our paper focuses on FPV gaze and actions, a joint model of gaze, hands and actions can be explored in future works.}

\noindent\textbf{Visualizing Egocentric Gaze for Action Recognition}. Moving beyond the accuracy, we provide more results to help understand our model. Specifically, we visualize the outputs of gaze estimation and action labels from our model, as well as the ground-truth gaze in Figure~\ref{fig:gaze_action}. Our gaze outputs often attend to foreground objects that the person is interacting with. We believe this is why the model is able to better recognize egocentric actions. Moreover, we found these visualizations helpful for diagnosing the error in action recognition. A good example is the second failure case in the third row (middle) of Figure~\ref{fig:gaze_action}, where our model successfully recognize the object as ``condiment container'' yet fail to distinguish the verb (``take'' vs.\ ``open''). Another example is the first failure case in Figure~\ref{fig:gaze_action}, where the recognition model is confused due to the appearance similarity between ``onion'' and ``cucumber''.

\subsection{Extension to EPIC-Kitchens Dataset}
Finally, we evaluate our model on EPIC-Kitchens dataset--the largest benchmark for FPV action recognition. By using a single network on the RGB frames, our model achieves state-of-the-art performance on EPIC-Kitchens. \newedit{Our results were ranked the 2nd for unseen environments, and the 4th for seen environments at EPIC-Kitchens Challenge 2020.}

\noindent \textbf{Dataset and Metric}. EPIC-Kitchens dataset has $39,594$ action instances. Similar to our dataset, each instance is described by the combination of a single verb and a single noun. There are in total 125 verb classes and 331 noun classes. We report results on both seen (s1) and unseen (s2) test sets from~\cite{Damen2018EPICKITCHENS}. Our results are evaluated by the server provided by~\cite{Damen2018EPICKITCHENS}. The main metric includes the top-1/5 accuracy for verb, noun and action labels.

\begin{table}[t]
\footnotesize 
\caption{\textbf{Action recognition results on Epic-Kitchens test sets}. We follow~\cite{Damen2018EPICKITCHENS} to report top1/top5 accuracy for verb / noun / action on seen (s1) and unseen sets (s2). Our results are further compared against previous methods that use a single model. At the time of this submission, our model (Ours+CSN152) ranks 1st for the unseen set and 3rd for the seen set on the EPIC-Kitchens Action Recognition Challenge Leaderboard.}
\centering
{
\tablestyle{2pt}{1.0}
\setlength{\tabcolsep}{3.5pt} 
\renewcommand{\arraystretch}{1.08} % Default value: 1
\begin{tabular}{c|c|ccc}
\multicolumn{1}{c}{\multirow{2}{*}{}}  
&\multicolumn{1}{c|}{\multirow{2}{*}{Method}}                                        
& \multicolumn{3}{c}{Top1/Top5 Accuracy} \\
\multicolumn{2}{c|}{}   & Verb & Noun &Action         \\ \hline 
\multirow{8}{*}{s1}  
&2SCNN~\cite{Damen2018EPICKITCHENS}   & 40.44 / 83.04 & 30.46 / 57.05 &13.67 / 33.25 \\
&TSN (fusion)~\cite{Damen2018EPICKITCHENS}     & 48.23 / 84.09 & 36.71 / 62.32 &20.54 / 39.79  \\
&LSTA-2S~\cite{sudhakaran2019lsta}      & 62.12 / 87.95 & 40.41 / 64.47 & 32.60 / 52.85 \\
&LFB Max~\cite{wu2019long} & 60.00 / 88.40 & 45.00 / 71.80 & 32.70 / 55.30 \\
&EPIC-Fusion~\cite{kazakos2019epic}   & 64.75 / 90.70 & 46.03 / 71.34 & 34.80 / 56.65 \\
&R(2+1)D~\cite{ghadiyaram2019large}   & 65.20 / 87.40 & 45.10 / 67.80 & 34.50 / 53.80 \\
&2SI3D+Obj~\cite{wang2019baidu}     & \textbf{69.80} / \textbf{90.95} & \textbf{52.27} / \textbf{76.71} & \textbf{41.37} / \textbf{63.59} \\ 
&Ours      & 68.51 / 89.32 & 49.96 / 72.30 & 38.75 / 59.00  \\ \hline
\multirow{8}{*}{s2}  
&2SCNN~\cite{Damen2018EPICKITCHENS}   & 36.16 / 71.97 & 18.03 / 38.41 & 7.31 / 19.49 \\
&TSN (fusion)~\cite{Damen2018EPICKITCHENS}      & 39.40 / 74.29 & 22.70 / 45.72 & 10.89 / 25.26 \\
&LSTA-2S~\cite{sudhakaran2019lsta}      & 48.89 / 77.88 & 24.27 / 46.06 & 18.71 / 33.77 \\
&LFB Max~\cite{wu2019long} & 50.90 / 77.60 & 31.50 / 57.80 & 21.20 / 39.40 \\
&EPIC-Fusion~\cite{kazakos2019epic}   & 52.69 / 79.93 & 27.86 / 53.78 & 19.06 / 39.40\\
&R(2+1)D~\cite{ghadiyaram2019large}   & 57.30 / 81.10 & 35.70 / 58.70 & 25.60 / 42.70 \\
&2SI3D+Obj~\cite{wang2019baidu}     & 58.96 / \textbf{82.69} & 33.90 / 62.27 & 25.20 / \textbf{45.48} \\ 
&Ours      & \textbf{60.05} / 81.97  & \textbf{38.14} / \textbf{63.81}  & \textbf{27.35} / 45.24 
\end{tabular}}\vspace{-1em}
\label{table:Epicaction}
\end{table}

\begin{table}[t]
\footnotesize 
\caption{\newedit{\textbf{Ablation results on Epic-Kitchens}. We report top1/top5 accuracy for verb / noun / action on seen (s1) and unseen sets (s2).}}
\centering
{
\tablestyle{2pt}{1.0}
\setlength{\tabcolsep}{3.5pt} 
\renewcommand{\arraystretch}{1.08} % Default value: 1
\begin{tabular}{c|c|ccc}
\multicolumn{1}{c}{\multirow{2}{*}{}}  
&\multicolumn{1}{c|}{\multirow{2}{*}{\newedit{Method}}}                                        
& \multicolumn{3}{c}{\newedit{Top1/Top5 Accuracy}} \\
\multicolumn{2}{c|}{}   & \newedit{Verb} & \newedit{Noun} & \newedit{Action}  \\ \hline 
\multirow{3}{*}{\newedit{s1}}  
&\newedit{I3D-Res50}    & \newedit{56.37 / 85.93} & \newedit{37.60 / 67.38} & \newedit{24.54 / 45.35}  \\ 
%&Ours + I3D-Res50   & 56.48 / 85.92   & 37.98 / 64.16 & 25.07 / 45.15  \\ 
&\newedit{CSN152}      & \newedit{67.97 / 88.88} & \newedit{49.73 / 72.89} & \newedit{38.49 / 58.05}  \\ 
&\newedit{Ours + CSN152}      & \newedit{68.51 / 89.32} & \newedit{49.96 / 72.30} & \newedit{38.75 / 59.00}  \\ \hline
\multirow{3}{*}{\newedit{s2}}  
&\newedit{I3D-Res50}     & \newedit{46.87 / 77.97}  & \newedit{25.03 / 51.81}  & \newedit{15.03 / 32.43} \\
%&Ours + I3D-Res50      &47.15 / 77.50 & 25.20 / 52.13   & 15.23 / 31.75  \\
&\newedit{CSN152}      & \newedit{59.76 / 81.50}  & \newedit{37.18 / 63.84}  & \newedit{26.90 / 44.45}  \\
&\newedit{Ours + CSN152}      & \newedit{60.05 / 81.97}  & \newedit{38.14 / 63.81}  & \newedit{27.35 / 45.24} 
\end{tabular}}\vspace{-1em}
\label{table:epic_ablation}
\end{table}

\noindent \textbf{Training Details}. We followed a similar training protocol as EGTEA Gaze+ with some modifications. \newedit{For all our experiments on EPIC-Kitchens, we restrict our models to use RGB frames only, as computing and storing optical flows for this dataset is expensive.} First, gaze tracking data is not available on EPIC-Kitchens dataset, hence we replaced the gaze prior with uniform distribution. Second, to further improve the performance, we replaced the I3D backbone with a more recent CSN152 network pre-trained by~\cite{tran2019video}. Third, by using a heavy backbone, we only considered a single RGB stream and reduced the batch size to 16. Finally, we found it helpful to prevent over-fitting by early stopping the training. Specifically, we only trained the network for 18 epochs where the learning rate was decayed by 1/10 at epoch 15. This strategy helps to boost the performance on both seen and unseen test set, and has a larger performance impact on the unseen set.

\noindent \textbf{Results and Discussion}. Our results are presented in Table~\ref{table:Epicaction} and compared to state-of-the-art methods~\cite{sudhakaran2019lsta, wu2019long, kazakos2019epic, ghadiyaram2019large, wang2019baidu}. Our final model, using only RGB frames, achieves state-of-the-art results in comparison to all prior work, including those use optical flow~\cite{sudhakaran2019lsta}, object detector~\cite{wang2019baidu} or audio data~\cite{kazakos2019epic}. Our single model outperforms the best published single-model entries~\cite{kazakos2019epic, ghadiyaram2019large} by a significant margin of 4.0\%, 1.8\% on the seen and unseen set, respectively. Moreover, our single model even beats large ensembles~\cite{kazakos2019epic}. \newedit{During the 2020 challenge, our single model results were ranked the 2nd on the unseen test set, and the 4th on the seen test set. Other top ranked entries used object detector and model ensembles~\cite{wang2019baidu}. We point out that the uniform gaze prior used in our model is perhaps over-simplified, yet our model still achieves very competitive results on this challenging benchmark. We speculate that better modeling of the gaze prior, e.g., using hand locations, might further improve the performance.}

\newedit{\noindent \textbf{Ablation Study}. To better understand different components of our model, we present a further ablation study on EPIC-Kitchens. To this end, we consider using two backbones networks: I3D-Res50 from~\cite{wang2018none} pre-trained on Kinetics~\cite{carreira2017quo} and CSN152 from~\cite{tran2019video} pre-trained on IG-65M~\cite{tran2019video}. Both backbones use only RGB frames. We then attached our model to CSN152 --- our model reported on the EPIC-Kitchens Challenge 2020. As a point of reference, I3D-Res50 has a mean class accuracy of $54.6\%$ on the first split of EGTEA, similar to two-stream I3D joint baseline (27 layers with an accuracy of $54.2\%$). The major performance boost comes from the heavy backbone and pre-training. Using CSN152 alone improves the results of I3D-ResNet50 by a large margin. The top-1/5 action accuracy increases by +$14.0\%$ / $12.7\%$ on seen set (s1) and by +$11.9\%$ / $12.0\%$ on unseen set (s2). On top of the strong CSN152, our model further improves the results by a noticeable margin (+$0.3\%$ / $1.0\%$ on seen set (s1) and by +$0.5\%$ / $0.8\%$ on unseen set (s2) for top-1/5 action accuracy). This trend is consistent with our results on EGTEA.}

%% file: conclusion.tex
\section{Conclusion and Future Work}
\label{sec:conclusion}
In this paper, we considered the task of joint gaze estimation and action recognition in first person video. To facilitate our research, we introduced a new dataset---EGTEA Gaze+. Our dataset comes with video recording, gaze tracking and hand masks, thereby providing the most comprehensive benchmark to understand egocentric gaze and actions. Moving beyond the dataset, we presented a novel deep model that can jointly estimate FPV gaze and recognition FPV actions. At the core of our model lies in the innovation of probabilistic modeling of human gaze using a deep model. We evaluated our model on EGTEA Gaze+ and demonstrated its superior performance. More importantly, the same model was applied to the largest FPV action recognition benchmark (EPIC-Kitchens) and achieved state-of-the-art results. We believe our dataset and model offer new insights for connecting egocentric gaze and actions, and thus providing a solid step towards advancing First Person Vision.